\documentclass[10pt,twocolumn]{article}
\usepackage{amsmath,amssymb}
\usepackage{graphicx}
\usepackage{subfigure}
\usepackage{color}
\usepackage{verbatim}
\usepackage{gensymb}
\usepackage[numbers,sort&compress]{natbib}
\usepackage{authblk}

\begin{document}

\title{Mouse Pose Estimation From Depth Images}

\author[1]{Ashwin~Nanjappa}
\author[1]{Li~Cheng\thanks{Corresponding author with email address: chengli@bii.a-star.edu.sg \\ The project is partially funded by A*STAR JCO1431AFG120.}}
\author[1]{Wei~Gao}
\author[1]{Chi~Xu}
\author[2]{Adam Claridge-Chang}
\author[3]{Zoe~Bichler}
\affil[1]{Bioinformatics Institute, A*STAR, Singapore}
\affil[2]{Institute of Molecular and Cell Biology, A*STAR, Singapore}
\affil[3]{National Neuroscience Institute, Singapore}

\date{}
%




\maketitle

\begin{abstract}
We focus on the challenging problem of efficient mouse 3D pose estimation based on static images, and especially single depth images.
We introduce an approach to discriminatively train the split nodes of trees in random forest to improve their performance on estimation of 3D joint positions of mouse.
Our algorithm is capable of working with different types of rodents and with different types of depth cameras and imaging setups.
In particular, it is demonstrated in this paper that when a top-mounted depth camera is combined with a bottom-mounted color camera, the final system is capable of delivering full-body pose estimation including four limbs and the paws.
Empirical examinations on synthesized and real-world depth images confirm the applicability of our approach on mouse pose estimation, as well as the closely related task of part-based labeling of mouse.
\end{abstract}

\section{Introduction}

The study of mouse behavior, referred to as behavioural phenotyping,
is an important topic in neuroscience~\cite{SchCla:con12}, pharmacology~\cite{GiaSeg:bjp10} and other related fields in biomedical sciences.
For example, it plays a key role in studying neurodegenerative diseases~\cite{SteEtAl:pnas07},
is used in modeling human psychiatric disorders~\cite{CryHol:nrdd05} and aids in better understanding of the genetics of brain disorders~\cite{Abb:nature13},
due to the known homology between animals and humans.
Automated analysis of mouse behavior has the potential to improve reproducibility and to enable
new kinds of psychological and physiological experiments~\cite{SchCla:con12}, besides efficiency and cost concerns.
From a computer vision standpoint, it presents exciting challenges including tracking~\cite{BraBel:cvpr05},
activity recognition~\cite{JhuEtAl:naturecomm10}, and social behavior analysis~\cite{BurEtAl:cvpr12}, based on color video feeds of mice behavior.
Existing efforts have mainly focused on analyzing two-dimensional color images.
On the other hand, recent growth of commodity depth cameras, like XBox Kinect, makes it possible to investigate mouse behavior further into three dimensions.
This opportunity motivates us to consider efficient 3D mouse pose estimation based on single depth images.
By pose estimation of a mouse, we refer to the collective estimation of the 3D positions of its major body joints, which is the fundamental problem in mouse behavior analysis.
Once the full 3D poses of mouse over time are obtained, behavioral and activity related information which are commonly used can be deduced in a rather straightforward manner.

Our work has four main contributions:
First, we introduce an approach to discriminatively train the split nodes of trees in a random forest, one node at a time, to improve their performance.
We adapt this approach to the regression task of joint position estimation of mouse and also to the classification task of body part labeling of mouse from depth images.
Second, based on our discriminatively trained random forest, we propose an efficient and effective system to estimate the 3D pose of a mouse based on single depth images.
To our best knowledge, this is the first such attempt to address this fundamental task.
Additionally, we also propose a method for the related task of labeling mouse body parts based on single depth images.
Third, we demonstrate the flexibility of our methods: it is designed to work with various depth cameras such as structured illumination or time-of-flight (ToF)
and with different imaging setups and with additional cameras, like a bottom-mounted color camera to facilitate the estimation of \emph{full-body} pose of the mouse.
Finally, we introduce a simple 3D mouse skeletal model, a mouse pose visualization and synthesis engine and a dedicated cage setup.
This is the first such kind of academic effort to our knowledge and might be useful for other researchers in this field.

\subsection{Related work}

\begin{figure*}[!tp]
\centering
\subfigure[]
{
    \includegraphics[width=.31\linewidth]{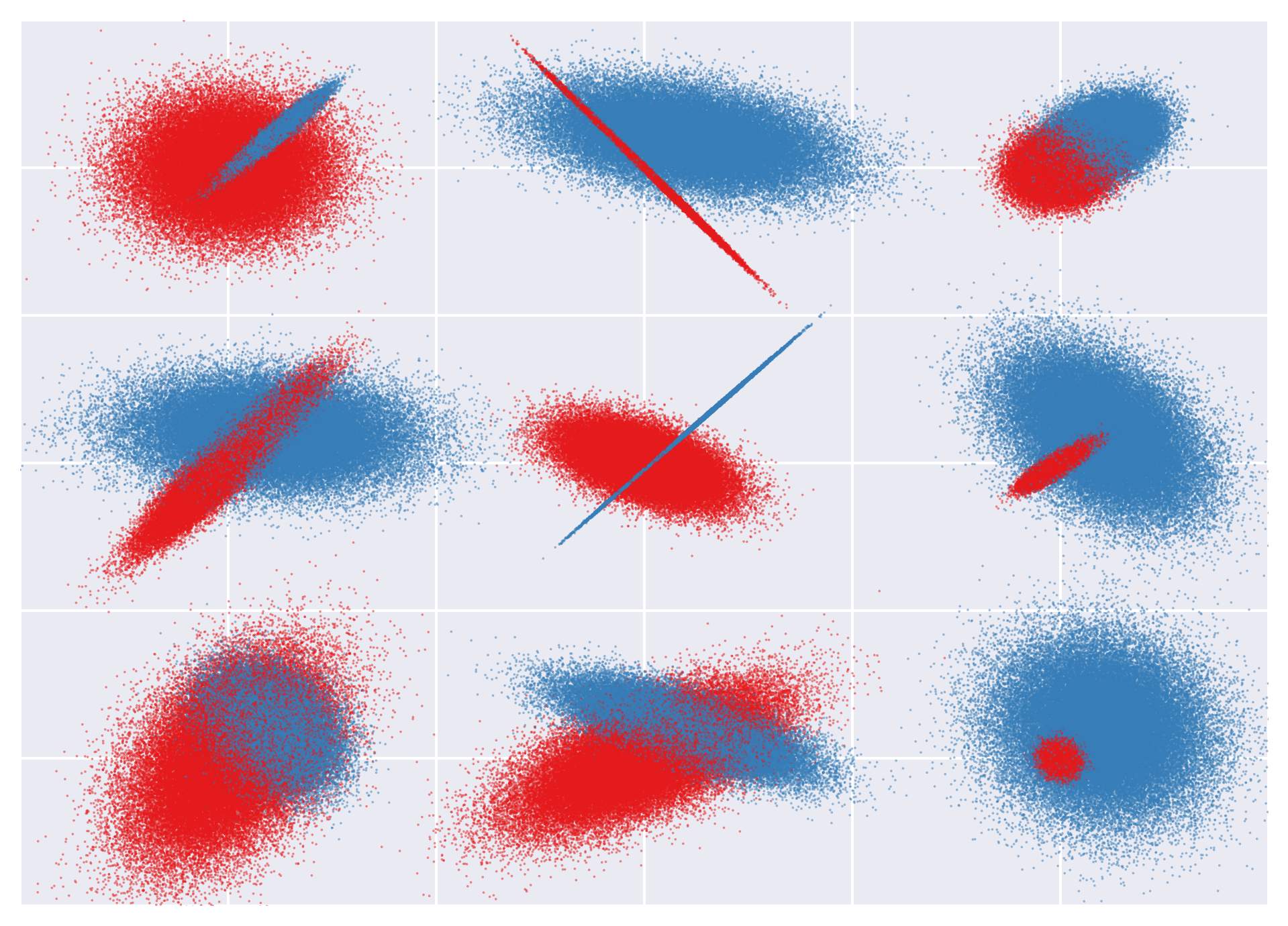}
    \label{fig:gauss}
}
\subfigure[]
{
    \includegraphics[width=.31\linewidth]{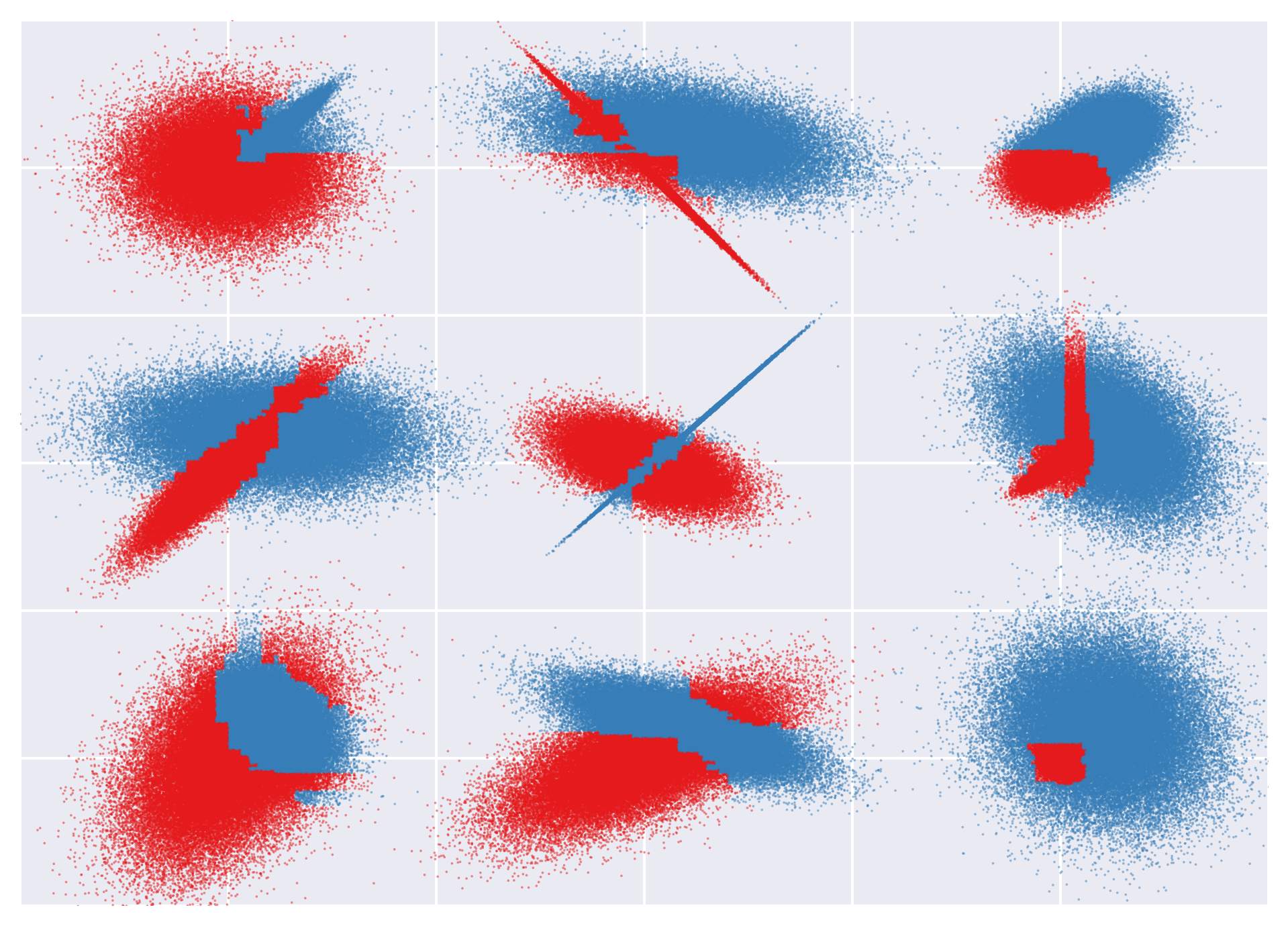}
    \label{fig:gauss_base}
}
\subfigure[]
{
    \includegraphics[width=.31\linewidth]{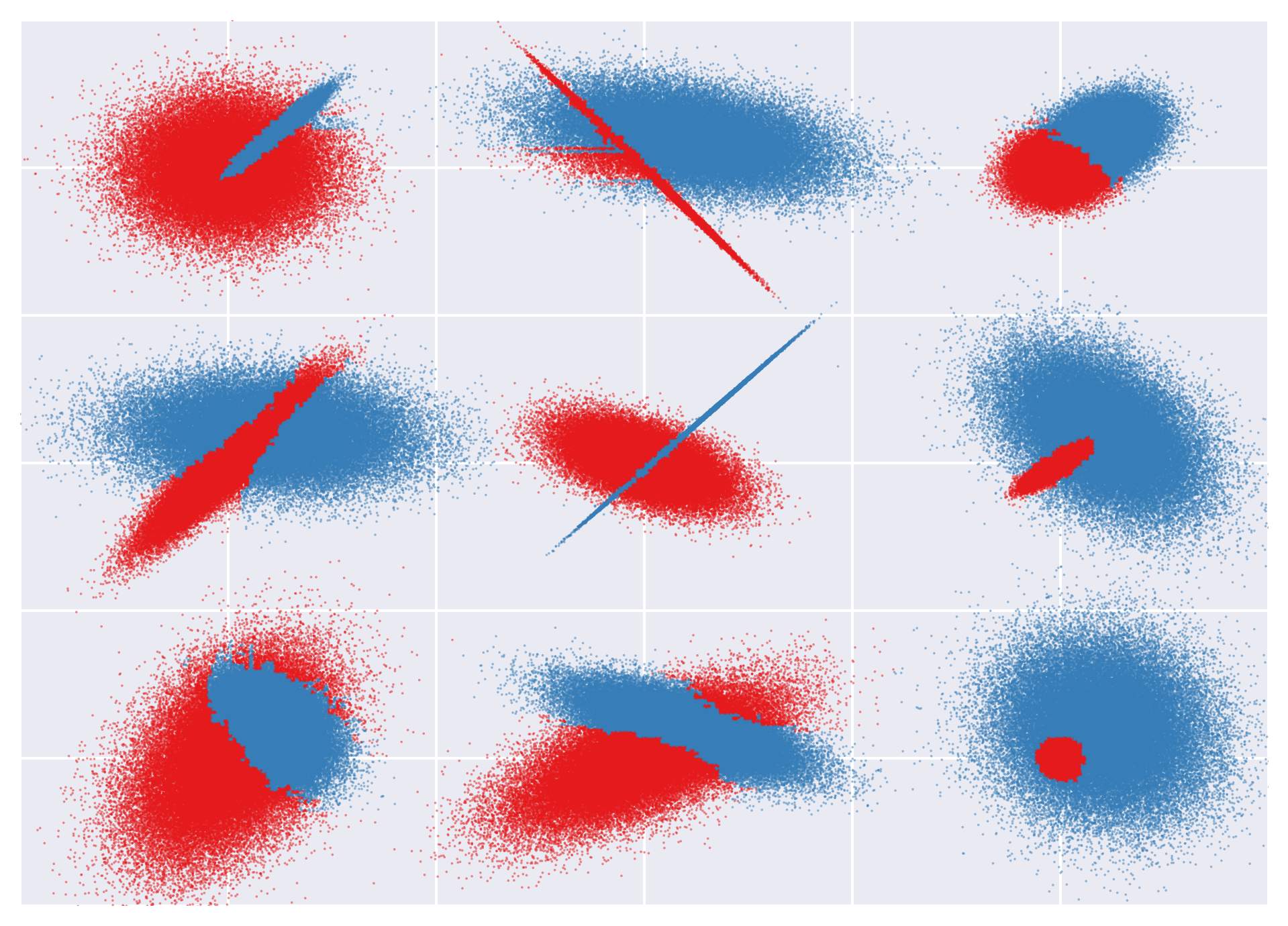}
    \label{fig:gauss_perturb}
}
\caption
{
    \footnotesize
    {
        (a) Points randomly sampled from 18 Gaussian distributions in 2D. Blue and red colors signify the two classes. Each distribution contributes to points of a single class.
        (b) shows the classification results of the base forest on a test dataset.
        (c) shows the classification results of the discriminatively trained forest on the same test dataset.
        It can be seen visually that the discriminatively trained forest can better differentiate between the two classes.
    }
}
\end{figure*}

There has been substantial progress in mouse tracking using color cameras.
Many commercial software used for studying social interaction of mice, such as HomeCageScan, EthoVision, AnyMaze, PhenoTracker
and MiceProfiler~\cite{ChaEtAl:nm12} use this technique to track up to two mice from color videos.
A common issue with tracking is that it is highly sensitive to small color or texture perturbations over time.
This often gives rise to failures, which makes a manual re-initialization of the system necessary and inevitable if used to track for long durations.
There has been some recent progress in long-duration tracking, such as~\cite{BurEtAl:cvpr12,WeiEtAl:nc13,OhaEtAl:jnm13}, which can work robustly without such manual intervention.
This is typically achieved with the help of invasive methods such as attaching RFID chips or applying dedicated fur dyes to maintain identities reliably over time.

In behavior phenotyping,
existing systems rely on discriminative learning methods such as SVM
and conditional random fields (CRFs)~\cite{WanEtAl:bmvc09,JhuEtAl:naturecomm10,BurEtAl:cvpr12} that directly focus on classifying behaviours or interactions.
These discriminative methods are usually supported by extracting visual features from 2D videos,
such as spatio-temporal interest points (STIP)~\cite{WanEtAl:bmvc09} obtained by convolution of video sequence with filters designed to respond
to salient points in the video volume, on which the visual bag-of-word features can be constructed.
One key drawback is that these approaches are based on highly sophisticated features that are usually inscrutable by behavioral neuroscientists.
This might lead to undesirable consequences when one tries to interpret the empirical findings, even if the classification performance is acceptable.
An alternative scheme is to instead rely on single-frame based pose estimation, which is a lot easier for domain experts to understand and interpret.

Meanwhile, the lack of three-dimensional information
hinders the ability of existing approaches to access rich spatial context that is crucial for elucidating mouse behaviors and interactions.
As a remedy, multi-camera setups have been considered~\cite{SheEtAl:plosone13}, which are relatively expensive, cumbersome, slow,
and the results are unreliable due to long-standing issues with multi-view vision such as sensitivity to textures, background clutters and lightings.

The recent advance of commodity depth imaging opens the door to drastic progress in this direction.
These depth cameras are currently based on either structured illumination or time-of-flight (ToF) technologies.
The most related work in the field~\cite{OuyEtAl:jnm11} is a low-res body pose estimation that involves the 3D location and orientation of the entire mouse based an overhead Primesense depth camera.
Our approach may be viewed as the next step to reveal high-resolution details of 3D mouse full-body skeleton.
Note the mice dataset of~\cite{BurEtAl:cvpr12} has also been studied in~\cite{BurEtAl:bmvc13} for a similar low-resolution pose estimation purpose in 2D based on gray-scale images.

%
%
Similar trends have been observed recently in computer vision problems related to humans, which have already demonstrated the unique value of utilizing depth cameras.
These include depth-image based pose estimation of the human body~\cite{ShoEtAl:pami13}, head~\cite{FanEtAl:ijcv13}, and hand~\cite{Xu2015Hand}.
Nonetheless, the set of challenges for mouse pose estimation are distinct and cannot be accomplished by a trivial re-implementation of these existing methods.
A lab mouse is noticeably smaller than a typical human hand, is highly deformable and is highly agile attaining over 1m/s maximum velocity~\cite{BonEtAl:ajpricp06} in a small enclosed space.
For any practical camera setup, occlusions of body parts such as limbs and paws of mice are common problems.
These factors also inspire us to consider dedicated cage and camera setups which will be described in later sections.

\section{Discriminative training of decision tree nodes}
\label{sec:dis_train}

In this section, we introduce a discriminative learning approach that improves the performance of random forests.
The basic idea is to discriminatively train the nodes of each tree $t \in \mathcal{T}$ of an existing random forest, modifying them to improve the performance on the input training dataset. The resulting forest is termed $\mathcal{T}_d$ to better differentiate from the original one.
A training dataset $D_d$, that is different from the training dataset $D_t$ used for creating $\mathcal{T}$, is generated separately for the purpose of this discriminative learning.
For each $t$ of $\mathcal{T}_d$, a randomly sampled subset $D'_d \subset D_d$ is applied to discriminatively train it.

The training process is applied iteratively on the nodes of $t$, a single node at a time, using $D'_d$ as input.
$D'_d$ is processed through $t$ by splitting it based on the tests stored at each split node in $t$.
For any $S \subseteq D'_d$ that arrives at a split node $q$, $S$ is split into $S_l$ and $S_r$ based on the test criteria stored in $q$.
Our discriminative process tries a random subset of $m$ tests in $q$, one at a time.
For each test a performance metric $\mathcal{E}(q, S) \rightarrow \mathbb{R}$ is used to measure the performance of the subtree rooted at $q$ for $S$.
The test delivering the best performance is stored at $q$ and $S$ is split further based on that test and this process is repeated on the rest of the nodes below $q$.
On the other hand if $q$ is a leaf node, then the estimated result that maximizes $\mathcal{E}(q, S)$ is stored at the leaf.

Proceeding iteratively in this manner, this discriminative training method perturbs nodes of $t$, while trying to maintain the existing structure of the tree.
There are two exceptional cases that are handled differently.
Let $l_n$ be the maximum number of training samples that are used to create a leaf node, and $|S|$ denotes the size of set $S$.
If $|S| \leq l_n$ at a split node $q$, then the subtree rooted at $q$ is replaced with a leaf node that maximizes $\mathcal{E}$.
Alternatively, if $|S| > l_n$ at a leaf node $q$, then a subtree is dynamically grown and replaces $q$.
Next we investigate the details of this discriminative training approach and analyze the effect of its tuning parameters using a simple example.

\subsection{A simple running example}

Consider points randomly sampled from two Gaussian distributions in $\mathbb{R}^2$, each representing a different class and let their spreads overlap each other.
As a simple binary classification problem, $10^6$ points are randomly sampled from nine such pairs of Gaussian distributions spread out in $\mathbb{R}^2$, as shown in Fig.~\ref{fig:gauss}.
Each point belongs to one of two classes and all points sampled from a distribution belong to the same class.
Mean positions $(\mu_x, \mu_y)$ such that $\mu \in (0, 1)$ and standard deviation $\sigma \in (0, 0.2)$ are used for the distributions.
Canonical training dataset $D_t$, discriminative training dataset $D_d$ and evaluation dataset $D_e$ are all sampled from these same distributions, with $|D_t|= |D_d|= |D_e|= 10^6$.

\subsection{Baseline method}

A random forest of classification trees is first trained using $D_t$ as the baseline forest.
Consider $S$ as the data points from the training set that arrive at a split node $q$.
A set of $m$ threshold real values $\Gamma = \{\gamma\}$ are uniformly sampled from the range $(0, 1)$ and each threshold $\gamma$ is used to split $S$ into two subsets $\{S_l, S_r\}$.
For nodes at even level, $\gamma$ splits $S$ horizontally and for nodes at odd level, $\gamma$ splits $S$ vertically.
To reduce the threshold space, we restrict the real value sampled to three decimal places.

The $\gamma$ that maximizes the gain $\mathcal{I}(\gamma)$ is chosen for the split node, where $\mathcal{I}$ is defined as:
\begin{align}
    \textstyle
\mathcal{I}(\gamma) = E(S) - \big( \frac { \left| S_l \right| }{ \left| S \right| } E(S_l)  + \frac { \left| S_r \right| }{ \left| S \right| } E(S_r) \big).
\label{eq:info_gain}
\end{align}
The entropy $E(S)$ is defined as:
\begin{align}
\textstyle
E(S) = - \sum\limits_{c \in \mathcal{C}} p(c) \log p(c),
\end{align}
where $p(c)$ is the normalized empirical histogram of labels for the data points in $S$, with $c$ indexing a particular class label and $\mathcal{C}$ the set of classes.

When $|S| < l_n$, then a leaf node is created which stores the class $c_l$:
\begin{align}
    c_l = \arg\max_{c \in \mathcal{C}} p(c).
    \label{eq:leaf_class}
\end{align}

A baseline classification forest is trained this way and evaluated using evaluation data $D_e$.
Each 2D point of $S$ arriving at a split node is split using the threshold $\gamma$ stored at the node, considered horizontal or vertical split depending on the level of the node.
When the point reaches a leaf node, it is assigned the class label $c_l$ stored there.
The final class label for a 2D point is picked from its $|\mathcal{T}|$ estimated class labels similar to Eq.~\eqref{eq:leaf_class}.

Fig.~\ref{fig:gauss_base} shows a visual result of classification of $10^6$ points in 2D by a baseline forest of $|\mathcal{T}| = 5$ where each tree has a maximum of $L = 20$ levels.
It can be compared to the ground-truth labels of the points shown in Fig.~\ref{fig:gauss}.
The classification accuracy, computed as the fraction of the points in the unseen test dataset that are classified correctly, is found to be $0.794$ for this experiment.

\subsection{Our discriminative training method}

\begin{figure}[!tp]
\centering
\includegraphics[width=.9\linewidth]{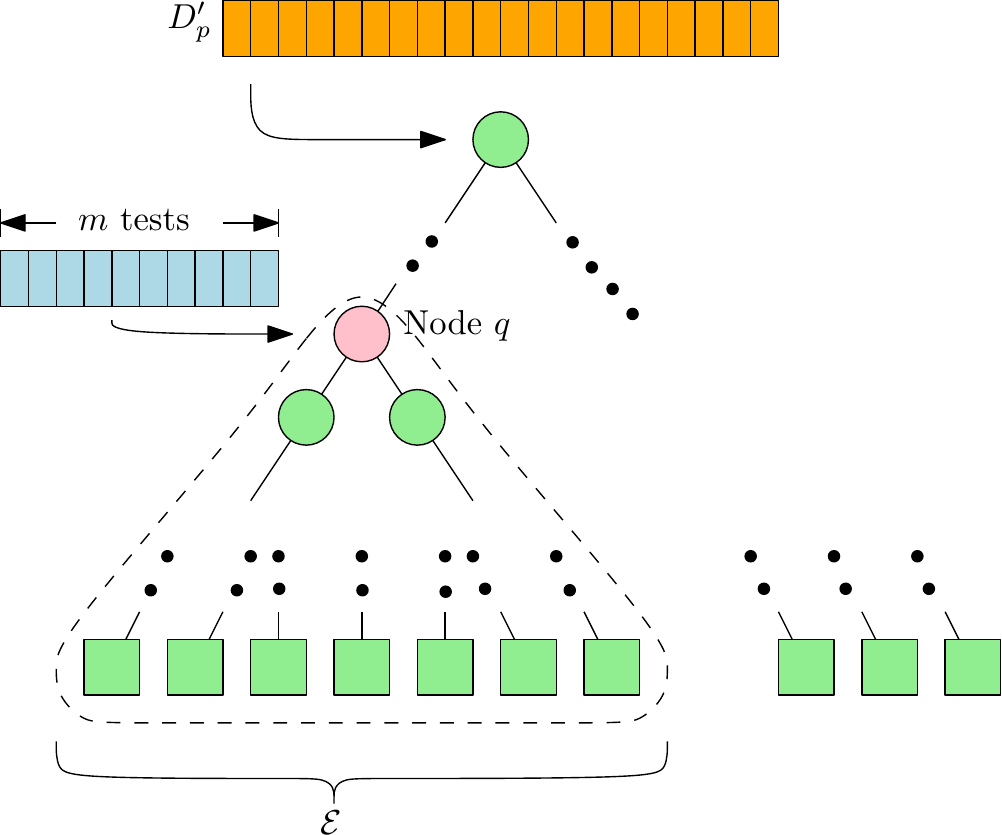}
\caption
{
    \footnotesize
    {
        Discriminative training of a node $q \in t$ using training data $D'_p$.
        A random set of $m_p$ tests is tried at $q$, one at a time, to minimize $\mathcal{E}(t, D'_p)$.
    }
}
\label{fig:dis_train}
\end{figure}

Next we take the baseline trees of $\mathcal{T}$ and iteratively train each of nodes of each tree as described earlier in a depth-first manner.
Specifically, at each split node $q$, a new set of $m$ thresholds is randomly sampled and subsequently examined using $\mathcal{E}$ at the related leaf nodes. 
Please refer to Eq.~\eqref{eq:Efunction} for the exact definition of $\mathcal{E}$ used here.

Fig.~\ref{fig:dis_train} illustrates this task at a split node $q$ from a baseline tree.
The nodes lying on the path from root node to $q$ have already been discriminatively learned.
Using the set of points $S$ that has reached $q$, a new set of $m$ thresholds is randomly sampled as before and the threshold that minimizes $\mathcal{E}$ is chosen and set in the split node.
This requires processing the points of $S$ through the nodes of the subtree rooted at $q$, as indicated in Fig.~\ref{fig:dis_train}.
The estimated class labels for $S$ are gathered from the leaves they reach in this subtree and $\mathcal{E}$ is computed for each threshold $\gamma$.

If $|S| \leq l_n$ at a split node $q$, then the entire subtree rooted at $q$ is replaced with a leaf, whose class is computed using Eq.~\eqref{eq:leaf_class}.
If $|S| > l_n$ at a leaf node, then the leaf is removed and a subtree is grown using baseline method and attached to the tree.
These steps allow dynamic modification of a baseline tree structure while training using $D'_d$.

Fig.~\ref{fig:gauss_perturb} shows a visual result of the classification of $D_e$ for such a discriminatively trained forest of $|\mathcal{T}_d| = 5$ trees, where the training is performed by scanning over all the tree nodes one time (i.e. one iteration).
By visual inspection, it can be easily noticed that this classification has finer classification boundaries that are closer to the ground-truth when compared to Fig.~\ref{fig:gauss_base}.
The classification accuracy is found to be $0.812$, a clear improvement over the baseline forest.

\subsection{Empirical studies of the running example}
In what follows, a series of experiments are conducted on the aforementioned simple running example to examine the performance changes when varying the internal parameters of our discriminative training process.

\subsubsection{Effect of discriminative training with second dataset}

We devise two experiments to ascertain the effectiveness of the combination of our discriminative training method with its second dataset $D_d$.
In the first attempt, we intend to find the effect of using a second dataset for the baseline method.
To do this, baseline trees are trained with the dataset $D_t \cup D_d$.
To ensure that trees of similar height as before are created, the value of $l_n$ is doubled.
This forest is found to have a classification accuracy of $0.796$.
This shows that merely applying more training data does not provide an on par improvement as with our discriminative training.

In the second attempt, we measure how much accuracy can be improved with discriminative training using $D_d$ with a small change: to use $\mathcal{I}(\gamma)$ instead of $\mathcal{E}$ on the leaf nodes as the internal error metric.
The classification accuracy of such a discriminatively trained forest on $D_e$ is found to be $0.804$, clearly better than the baseline, but not as accurate as using $\mathcal{E}$.
This shows that applying a second training dataset along with use of $\mathcal{E}$ that tests the performance of the entire subtree rooted at a node are both crucial for better accuracy of this method.
Thus it is this type of discriminative training that will be discussed in the rest of this paper.

\subsubsection{Effect of forest size}

\begin{figure}[!tp]
\centering
\includegraphics[width=.95\linewidth]{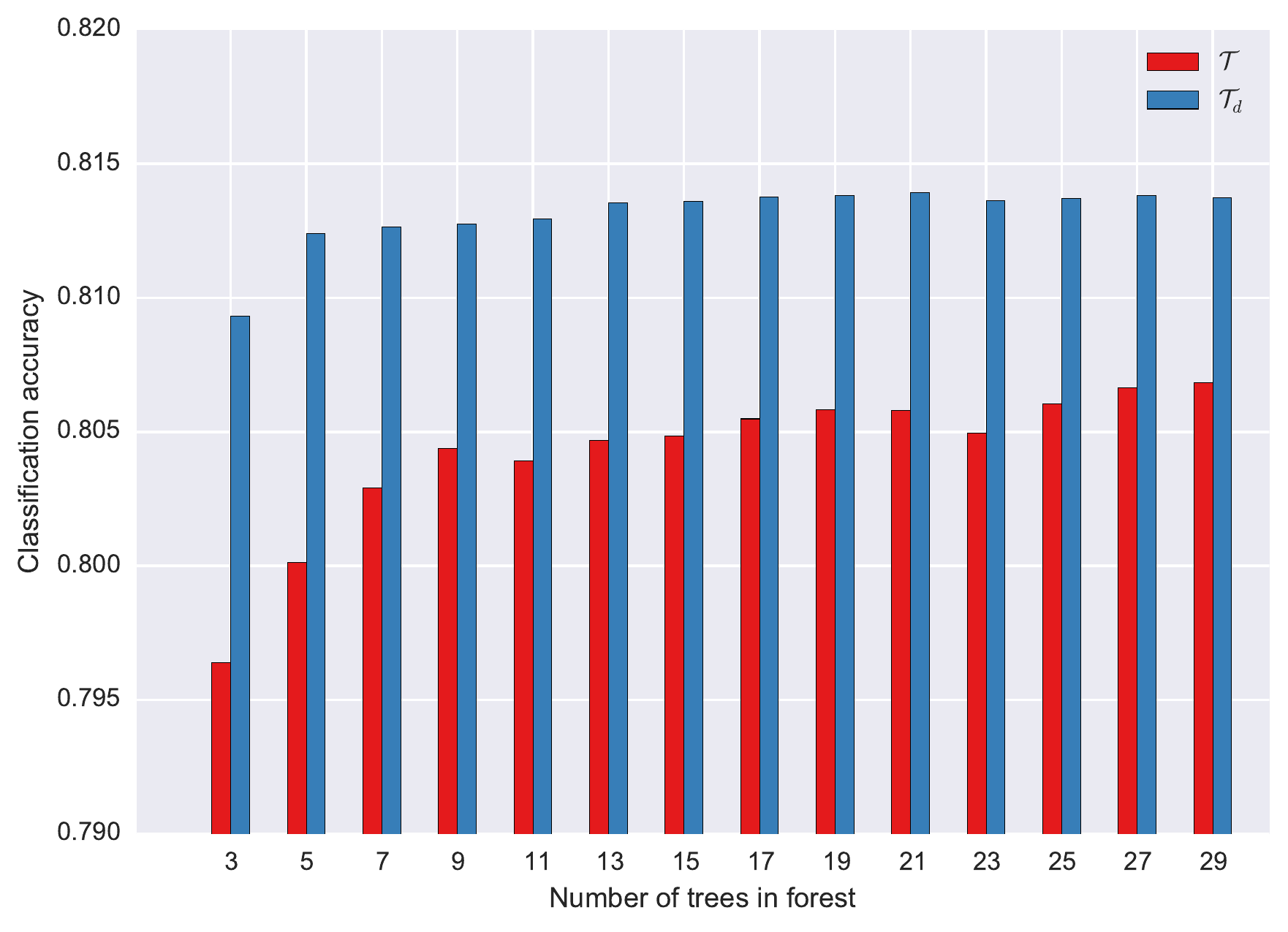}
\caption
{
    \footnotesize
    {
        Effect of forest size on classification accuracy of baseline and discriminatively trained forests.
    }
}
\label{fig:gauss_forest_size}
\end{figure}

In Fig.~\ref{fig:gauss_forest_size}, we investigate the effect of forest size on classification accuracy comparing both the baseline and its discriminatively trained forests.
It can be clearly seen that no matter how many trees are used, the discriminatively trained forest results in higher classification accuracy.
We observe that the baseline forest accuracy improves slowly with the number of trees, with the improvement lessening with increasing forest size.
The discriminatively trained forest accuracy remains almost stable after $|\mathcal{T}_d|=13$.
This also indicates that with discriminative training, the forest size can be considerably smaller without much loss in performance.

\subsubsection{Effect of $m$}
\label{sec:gauss_pool_size}

\begin{figure}[!tp]
\centering
\includegraphics[width=.95\linewidth]{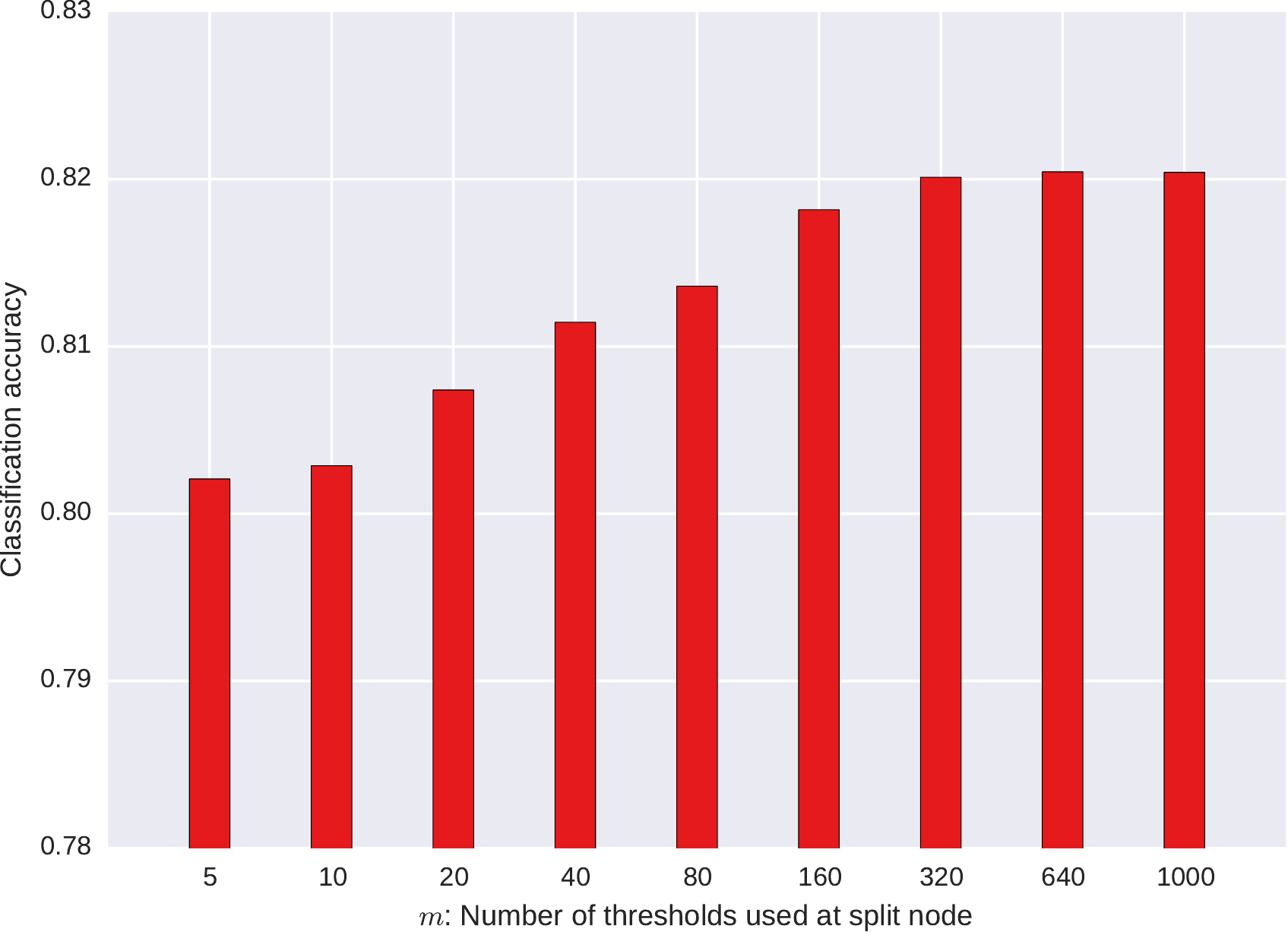}
\caption
{
    \footnotesize
    {
        Effect of feature size on classification accuracy.
    }
}
\label{fig:gauss_pool_size}
\end{figure}

In Fig.~\ref{fig:gauss_pool_size}, we investigate the effect of $m$, the number of thresholds picked for evaluation at each split node during discriminative training.
A baseline forest of $|\mathcal{T}_d| = 5$ and $m = 50$ is used for this experiment.
As mentioned earlier, we restrict the real value of the threshold to three decimal places, so there are $10^4$ thresholds to uniformly sample from.
We can observe that the classification accuracy increases with $m$, but levels away after $m = 320$ with an accuracy of $0.82$.
This may be attributed to the increasing correlation between the trees of a forest as $m$ increases.

\subsubsection{Effect of $l_n$}

\begin{figure}[!tp]
\centering
\includegraphics[width=.95\linewidth]{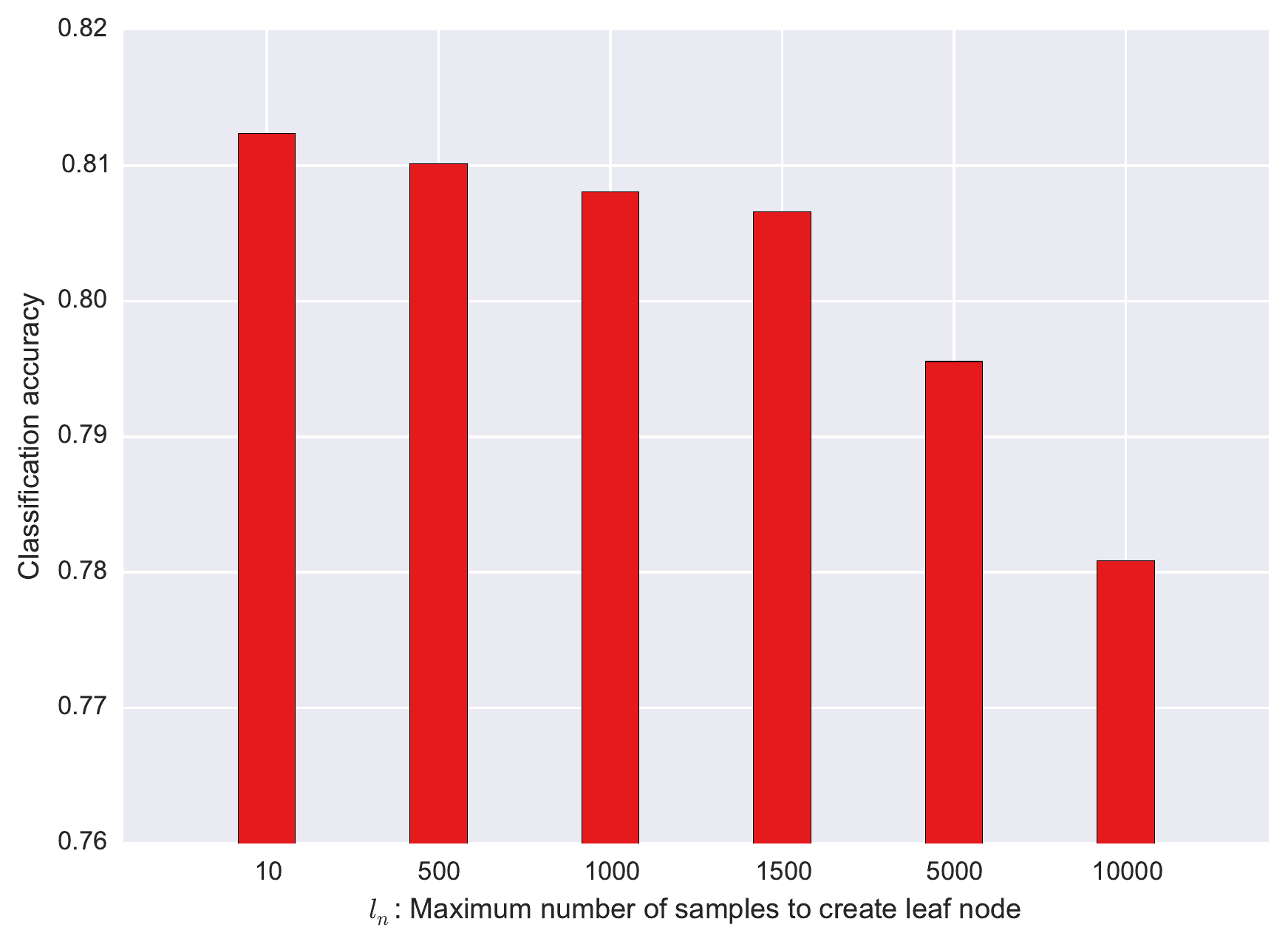}
\caption
{
    \footnotesize
    {
    Effect of maximum number of samples in leaf on classification accuracy.
    }
}
\label{fig:gauss_leaf_size}
\end{figure}

In Fig.~\ref{fig:gauss_leaf_size}, we look at the effect of $l_n$, the maximum number of data points required to create a leaf, on discriminative training.
The baseline forest is the same as used in Sec.~\ref{sec:gauss_pool_size}.
We can see that as $l_n$ increases, the classification accuracy decreases drastically, with accuracy decreasing to $0.78$ when $l_n = 10000$.
We note that the accuracy of discriminative training now decreases below that of its baseline forest.
This decrease in accuracy is attributed to the increasing number of subtrees of the baseline that undergo the shrink operation, being replaced with a leaf.

\subsubsection{Effect of node level}

\begin{figure}[!tp]
\centering
\includegraphics[width=.95\linewidth]{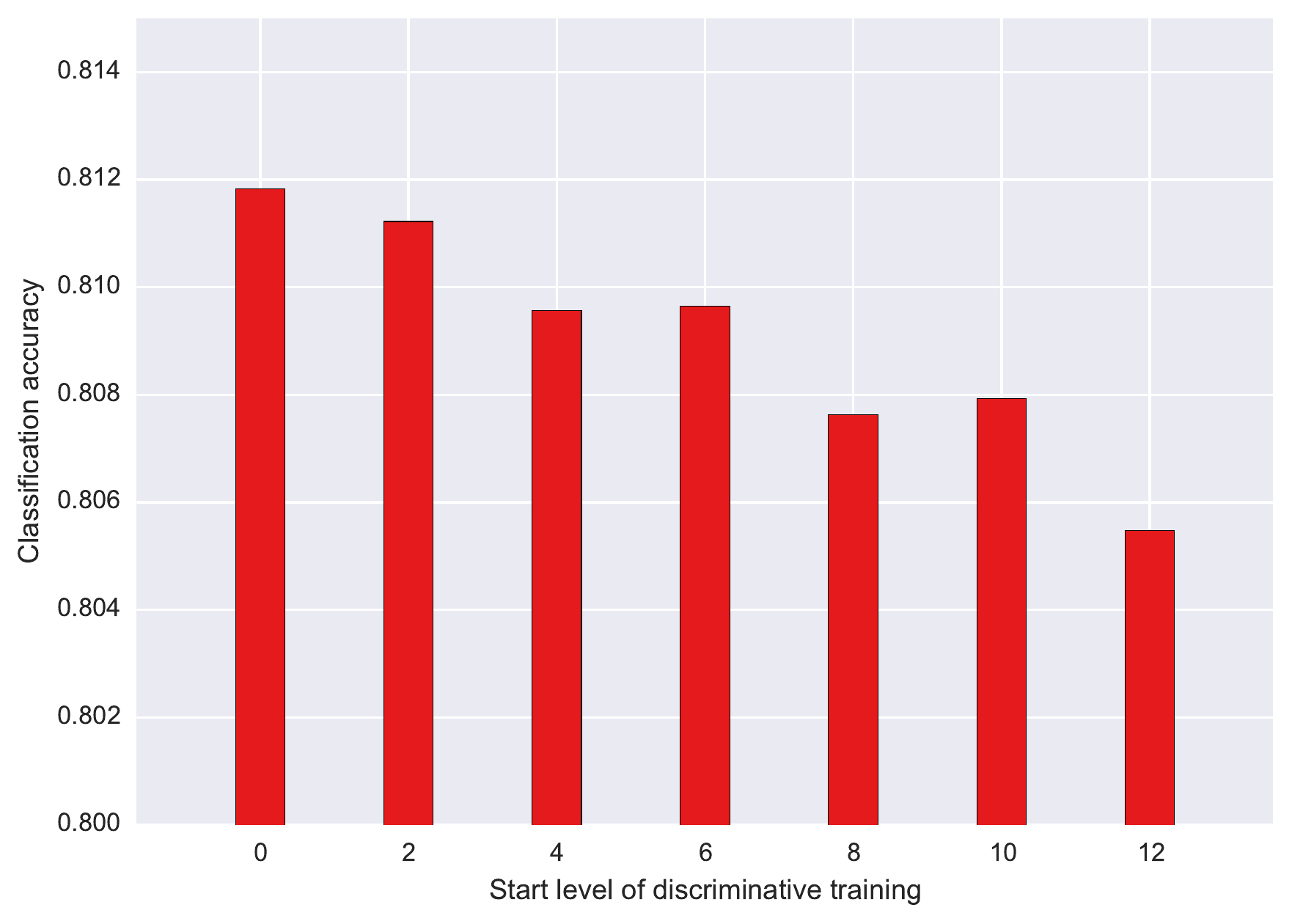}
\caption
{
    \footnotesize
    {
    Effect of start level for discriminative training on classification accuracy.
    }
}
\label{fig:gauss_start_level}
\end{figure}

In Fig.~\ref{fig:gauss_start_level}, we look at the effect of discriminative training of nodes at different levels on the classification accuracy.
In this experiment, we begin the process of discriminative training from nodes at a specified level in the tree.
When the starting level is $0$, this is equivalent to discriminative training applied on the full tree, which we have described earlier.
We increase the starting level of training in increments of $2$, since alternate level split nodes hold thresholds of X and Y axis respectively.
We see that as the starting level of discriminative training is lowered along the tree, the classification accuracy also decreases gradually.
Since every level $n$ typically has twice the number of nodes of level $n - 1$, this indicates that that the discriminative training at higher level nodes has a bigger effect per node than lower levels.

\subsubsection{Effect of multiple training iterations}

\begin{figure}[!tp]
\centering
\includegraphics[width=.95\linewidth]{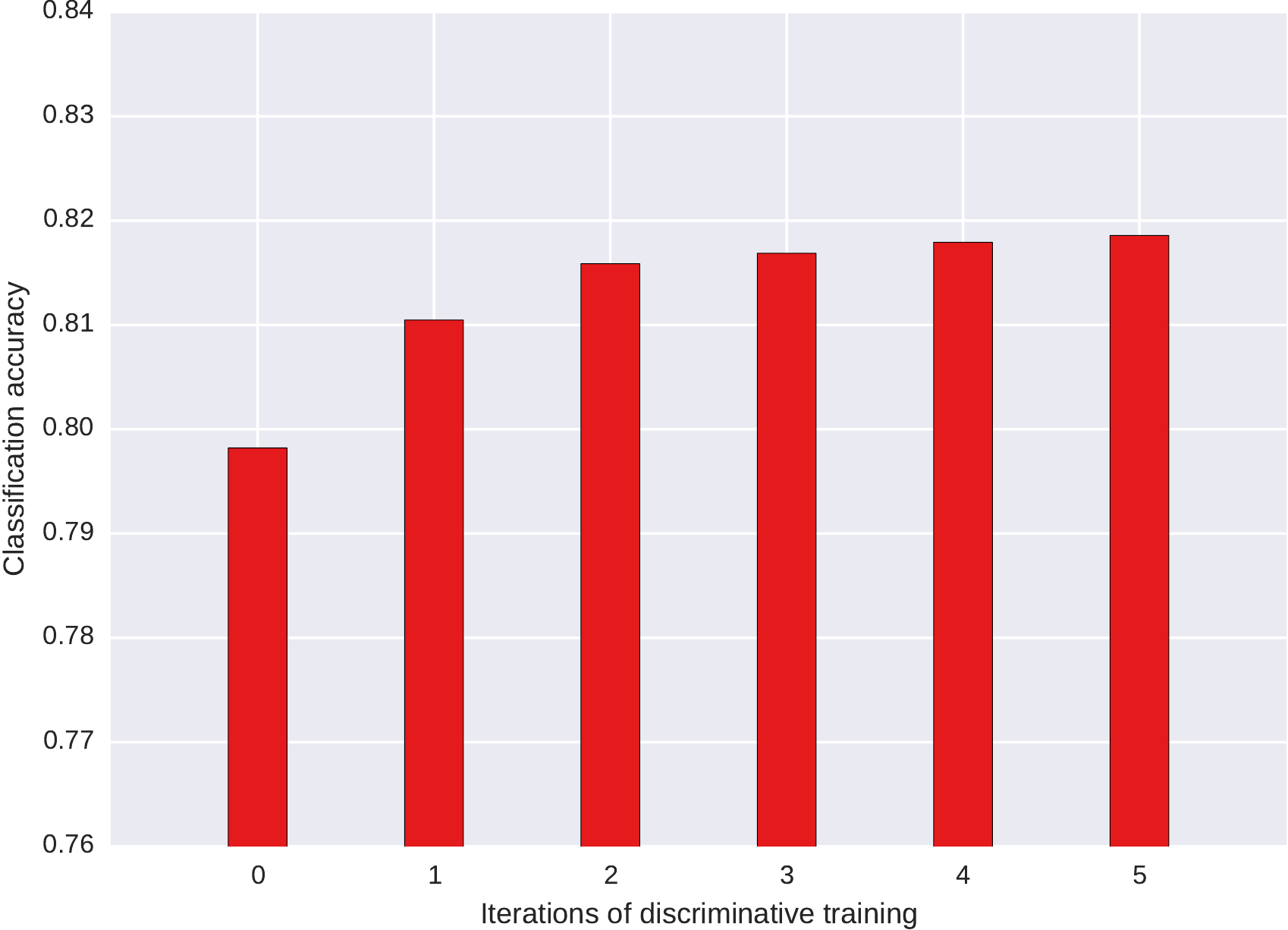}
\caption
{
    \footnotesize
    {
        Effect of multiple iterations of discriminative training on classification accuracy.
    }
}
\label{fig:gauss_iterations}
\end{figure}

In Fig.~\ref{fig:gauss_iterations}, we investigate into the effect of performing multiple iterations of discriminative training.
Iteration $0$ is the baseline forest and iteration $1$ applies discriminative training on the baseline with a new $D_p$ dataset.
In iteration $n = \{2, 3, 4, 5\}$, the discriminatively trained forest of iteration $n -1$ is used as the baseline and a new $D_p$ is created for training.
We observe that though the accuracy increase by repeated iterations of training is quite less after iteration $2$.
This is attributed to the tree structure and thresholds that stabilize after a few iterations and no further training helps after this.

In what follows, this discriminative training approach is applied to our regression forests for pose estimation and classification forests for part-based labeling of mouse from depth images as described in Sec.~\ref{sec:reg_split_node} and Sec.~\ref{sec:pbl_results} respectively.

\section{Overview of our mouse pose estimation system}

\begin{figure*}[!tp]
\centering
\includegraphics[width=.95\textwidth]{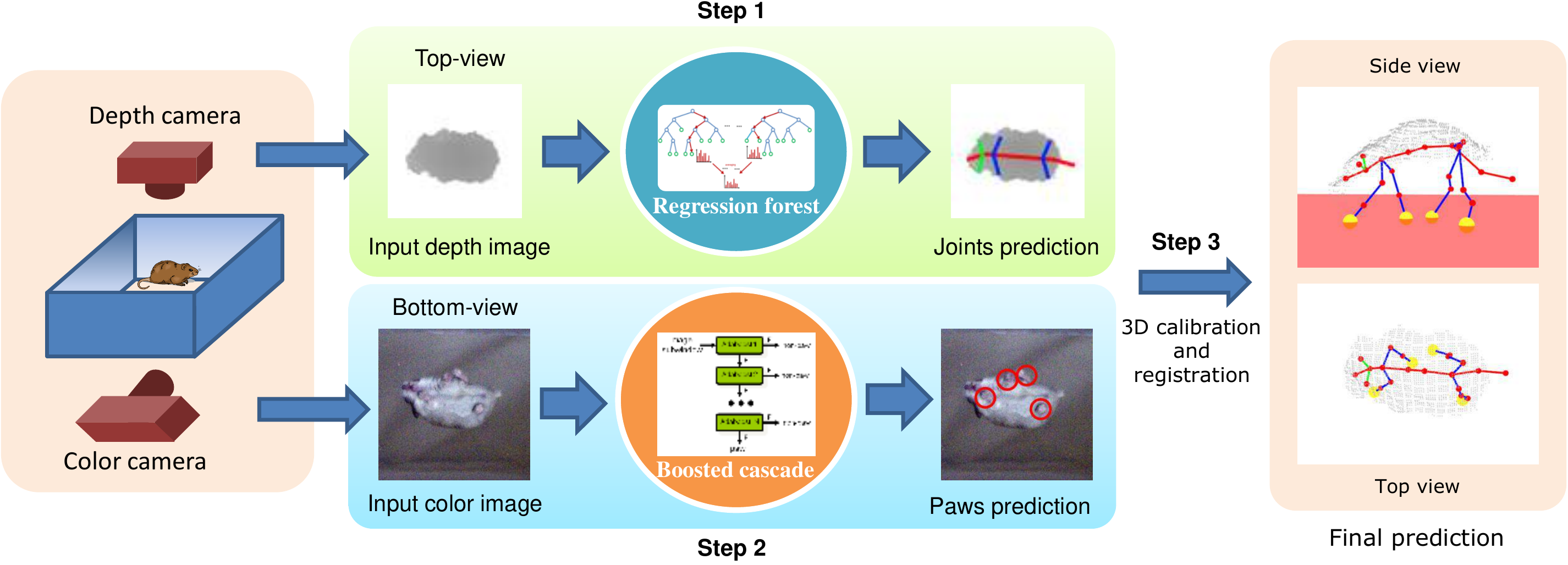}
\label{fig:pipeline}
\caption
{
    \footnotesize
    {
        The three steps of our approach:
        From top-view depth image, Step 1 estimates 3D positions of main body joints using a learned regression forest model.
        Step 2 uses a bottom-view color image to predict the locations of the four paws.
        Output of these two steps are fused in Step 3 to estimate the final full-body pose.
    }
}
\end{figure*}

Our system uses a setup of a top-mounted depth camera and a bottom-mounted color camera which delivers a synchronized pair of depth and color images.
Simple image preprocessing is used to segment the mouse patch from the depth image, as illustrated in the left image of step 1 in Fig.\ref{fig:pipeline}.
Step 1 of the system estimates the 3D locations of main body joints from a depth image using a regression forest.
It is difficult to uncover the limbs that are largely occluded from a top view.
To address this issue, step 2 takes as input a color image captured from below the glass floor of the cage, to determine locations of the paws by making use of a cascade detection procedure.
Finally, step 3 delivers the final full-body pose estimation by integrating these separate estimation results using 3D registration.

We next introduce our experimental setup and then our 3D mouse skeletal model and our 3D synthesis engine that generates 3D virtual mice.
Following that, we describe our base system and the application of discriminative training on it to reduce the error.

\subsection{Experimental setup}

\begin{figure}[!tp]
\centering
\includegraphics[width=.9\linewidth]{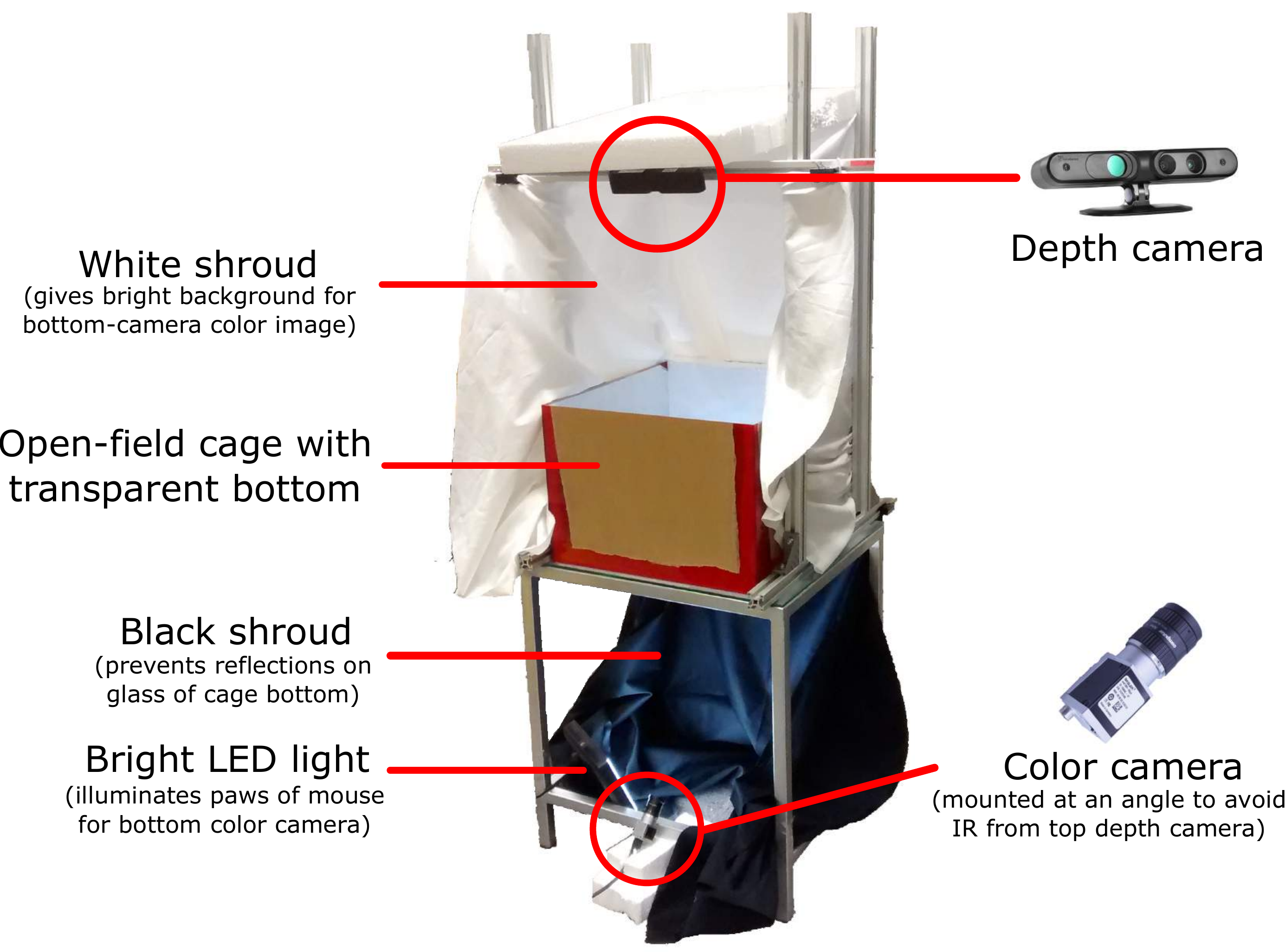}
\caption
{
    \footnotesize
    {
        Our experimental setup consisting of a cage, a top depth camera and a bottom RGB camera.
    }
}
\label{fig:setup}
\end{figure}

Our setup consists of a custom-built open-field apparatus (50cm$\times$45cm$\times$30cm) with transparent glass floor, a depth camera fixed $60$cm above and a color camera placed $60$cm below the cage, as shown in Fig.\ref{fig:setup}.
The bottom camera is mounted at a small angle deviating from the vertical axis to avoid direct incoming infrared light from the top-mounted depth camera.
In principle, the top-view depth image is used to capture the main body pose, while location of the paws are obtained from the bottom-view color image.
When working with both top- and bottom-mounted cameras, both are synchronized to ensure a pair of images is observed simultaneously at a time.

A grid of LEDs is placed near the bottom camera to illuminate the paws and improve the quality of images of this camera.
To prevent reflections appearing on glass floor bottom, a black cloth is used to encircle the bottom of the setup.
To prevent effect of external light sources, a white cloth is used to encircle the top of the setup.

\subsection{3D mouse model}

\begin{figure}[!tp]
    \centering
    \subfigure[]
    {
        \includegraphics[height=3cm]{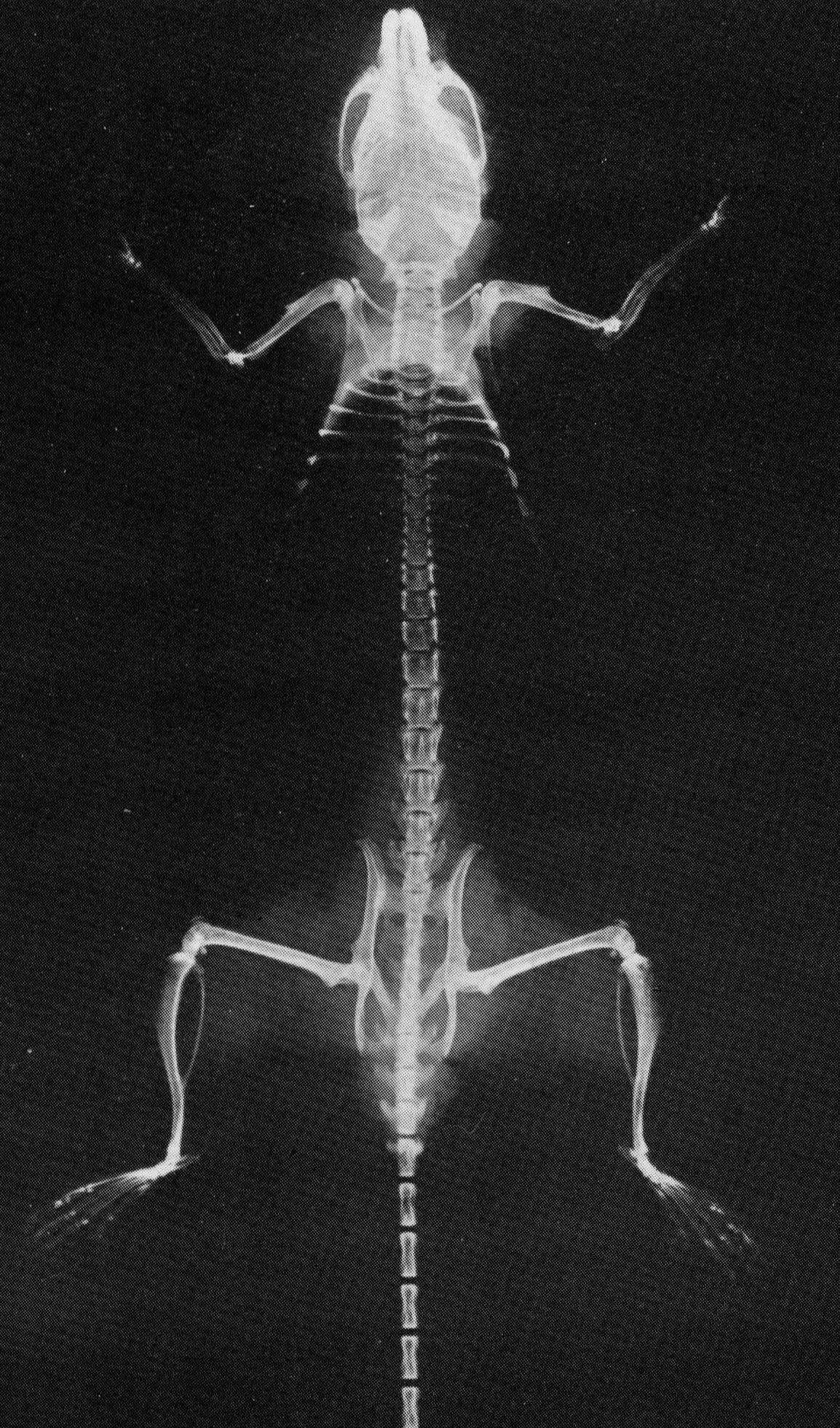}
        \label{fig:anatomy-a}
    }
    \subfigure[]
    {
        \includegraphics[height=3cm]{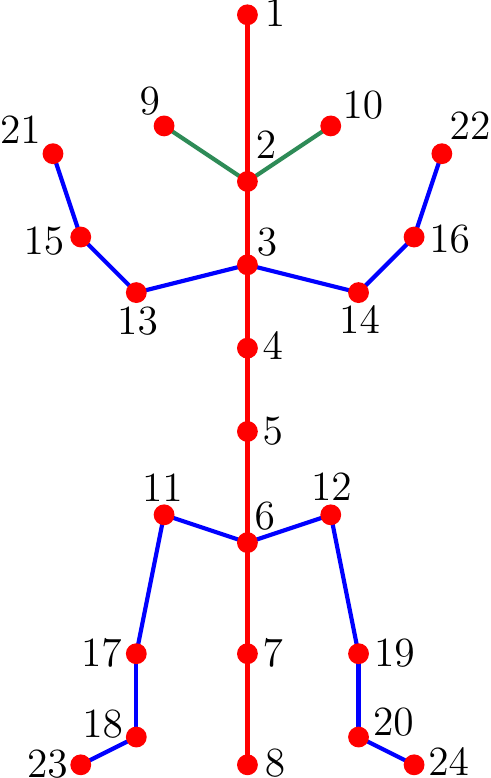}
        \label{fig:anatomy-b}
    }
    \subfigure[]
    {
        \includegraphics[height=3cm]{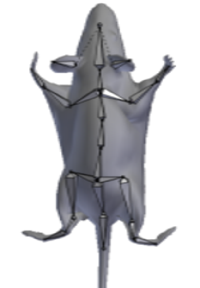}
        \label{fig:anatomy-c}
    }
    \subfigure[]
    {
        \includegraphics[height=3cm]{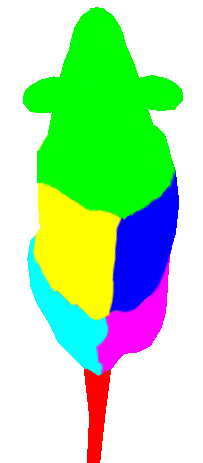}
        \label{fig:anatomy-d}
    }
\caption
{
    \footnotesize
    {
    (a) Skeletal anatomy of mouse.
    (b) Our kinematic model of mouse.
        The spine is colored in \textcolor{red}{red}, ears in \textcolor{green}{green} and the limbs and paws in \textcolor{blue}{blue}.
    (c) Our mouse model rendered with skin mesh.
    (d) Part-based labeling of our model with $6$ distinct colors: head in \textcolor{green}{green}, front-right in \textcolor{blue}{blue}, front-left in \textcolor{yellow}{yellow}, rear-right in \textcolor{magenta}{pink}, rear-left in \textcolor{cyan}{cyan} and tail in \textcolor{red}{red}.
    }
}
\end{figure}

We propose a 3D skeletal model of the mouse, containing 24 joints ($J=24$), as illustrated in Fig.\ref{fig:anatomy-b}.
It has been designed based on mouse skeletal anatomy~\cite{DogEtAl:pmb07,KhmEtAl:mib11} shown in Fig.\ref{fig:anatomy-a}, with few modifications:
(1) Joints: Mouse skeleton has more than 100 joints and we have simplified these to $24$ in our model.
(2) Ears: To help differentiate head and tail ends of a mouse, the ears are explicitly considered as two joints connected to main body.
(3) Tail: Two joints are assigned to approximately characterize the tail.
This approximation of tail is to account for the low resolution and noise of current consumer depth cameras, which
make it difficult to detect thin and long objects such as mouse tail in depth images.

To differentiate the limbs and the main body spine in visualization, bones are highlighted in unique colors.
Bones of the spine from head to tail are colored in \textcolor{red}{red}.
Bones of ear joints $\{9, 10\}$ are colored in \textcolor{green}{green}.
Bones connecting the rest of the joints 11-24 of the four limbs and paws are in \textcolor{blue}{blue}.
Fig.\ref{fig:anatomy-c} shows a 3D rendering of applying the proposed skeletal model with surface mesh and skin texture mapping.

\subsection{Synthesis engine}

\begin{figure}[!tp]
\centering
\includegraphics[width=0.9\linewidth]{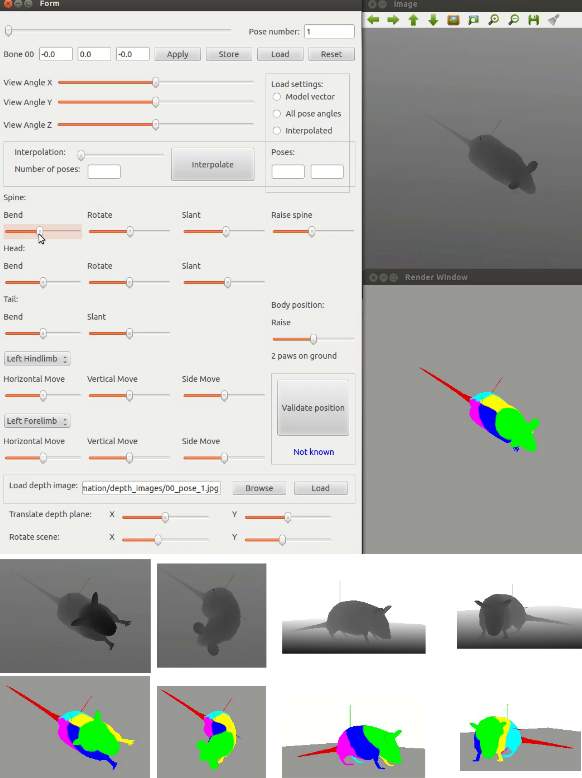}
\caption
{
    \footnotesize
    {
    Top: UI of our 3D virtual mouse engine, which is used to synthesize and visualize ground-truth mouse depth images.
    Bottom: Exemplar mouse poses (in part labels) and their rendered depth images.
    }
}
\label{fig:syn3DMouse}
\end{figure}

This skeletal model is animated to simulate most of the mouse full-body poses using a pose synthesis engine.
This is a GUI program we have developed to facilitate the easy visualization and modification of body joint angles and to efficiently generate sufficient amount of ground-truth synthetic depth images for training and testing purposes.
The GUI of this engine is shown at the top of Fig.\ref{fig:syn3DMouse}.
Using this engine and with the aid of an animation expert, 800 unique poses of typical mouse activity like standing, walking, running, turning and a combination of these poses are created.
A few exemplar poses and their depth images can be seen at the bottom of Fig.\ref{fig:syn3DMouse}.
Random pitch, roll and in-plane rotations and small variations in scale and bone lengths are applied to these poses to render the depth images for training.
To our knowledge, this is the first such reported academic effort in synthesizing and visualizing mouse poses based on a kinematic model.

\section{Our mouse pose estimation system}
\label{sec:reg_algo}

Now we are ready to present the three-step pipeline of our system as follows.

\subsection*{Step 1: Estimation of Main Body Joints from Top-view Depth Image}\label{sec:classifier}
\label{sec:test}

The main body joints 1-12 in Fig.\ref{fig:anatomy-b} consist of the joints of spine, ear and hip.
Starting with a set of simple depth features, this estimation is accomplished by developing a dedicated regression forest model that is influenced by the work of~\cite{Bre:ml01,ShoEtAl:pami13}.
In our context, a regression forest contains an ensemble of $|\mathcal{T}|$ binary decision trees.
Each tree $t \in \mathcal{T}$ is independently constructed by recursively growing the split nodes starting from the root split node,
and is trained on a dataset of synthesized depth images obtained by running our 3D mouse engine that renders the mesh of Fig.~\ref{fig:anatomy-c} in various poses, rotations and scales.

For depth features, a simple version of the popular depth features in~\cite{ShoEtAl:pami13,Xu2015Hand} are adapted.
At a given pixel 2D location $\mathbf{x} \in \mathbb{R}^2$ of an image $I$, denote its depth value as a mapping $d_I (\mathbf{x})$,
and construct a feature function $\phi$ by considering a 2D offset position $\mathbf{u}$ deviating from $\mathbf{x}$.
Following~\cite{Bre:ml01}, a binary test is defined as a pair of elements, $(\phi, \gamma)$, with $\phi$ being the feature function,
and $\gamma$ being a real-valued threshold.
When an instance $\mathbf{x}$ passes through a split node of our binary trees,
it will be sent to the left branch if $\phi(\mathbf{x})>\gamma$, and to the right branch otherwise.

\subsubsection*{Regression forest: Tree and Split Nodes}
\label{sec:reg_split_node}

Similar to existing regression forests in~\cite{ShoEtAl:pami13}, at a split node, we randomly select a small set of $m$ distinct features $\Phi := \{\phi_i\}_{i=1}^m$.
At every candidate feature dimension, a set of candidate thresholds $\Gamma$ is uniformly selected over the range defined by the empirical set of training examples in the node.
The best test $\big( \phi^*, \gamma^* \in \Gamma \big)$ is chosen from these features and accompanying thresholds, by maximizing the gain function defined next.
This procedure is then repeated until there are $L$ levels in the tree \textit{or} once the node contains fewer than $l_n$ training examples.

The split test is obtained by
\begin{align}
(\phi^*, \gamma^*)=\arg\max_{\phi \in \Phi, \gamma \in \Gamma} \mathcal{I} (\phi, \gamma),
\end{align}
where the gain $\mathcal{I}(\phi, \gamma)$ is defined as in Eq.\eqref{eq:info_gain}.
In our context, a training example refers to a pixel in the depth image, as well as its 3D location in the underlying 3D virtual mouse.
Therefore, any set of training examples $\hat{S}$ naturally corresponds to a point cloud in 3D space.
Denote $\mathbf{o}_{i \rightarrow j} \in \mathbb{R}^3$ the offset of example $i$ to the 3D location of joint $j$.
Let $\hat{S}_j :=\big\{ i \in \hat{S} \,\mid\, \|\mathbf{o}_{i \rightarrow j}\| < \epsilon \big\}$ be the subset of examples in $\hat{S}$ that are sufficiently close to joint $j$.
Denote $\bar{\mathbf{o}}_j := \frac{1}{|\hat{S_j}|} \sum_{i \in \hat{S}_j} \mathbf{o}_{i \rightarrow j}$ the mean offset of the set to $j$-th joint location.
The function $E$ is defined over a set of examples that evaluates the sum of deviations from the mean joint estimations:
\begin{align}
    \textstyle
E(\hat{S})= \sum_{j=1}^J \sum_{i \in \hat{S}_j} \big\| \mathbf{o}_{i \rightarrow j} - \bar{\mathbf{o}}_j \big\|_2,
\label{eq:entropy}
\end{align}
with the set of main body joints visible from top-mounted depth camera.
For a point cloud $\hat{S}$, a smaller $E(\hat{S})$ suggests a more compact cluster.
In this context, for $S \subseteq D'_d$ that arrives at a node $q$ having test $(\phi_q, \gamma_q)$, a set of $m$ tests $(\phi, \gamma)$ is attempted to maximize $\mathcal{E}_{(\phi, \gamma)}(t, D'_d)$.
$\mathcal{E}$ is adapted to regression as the reciprocal of the joint estimation error for the samples in $S$:
\begin{align}
    \textstyle
    \mathcal{E}(t, D'_p) = (\sum_{i \in D'_p} \sum_{j = 1}^J \| \hat{\mathbf{p}}_j - \mathbf{p}_j \|_2)^{-1},
    \label{eq:reg_error_metric}
\end{align}
where the Euclidean distance between true position $\mathbf{p}_j$ and the estimated position $\hat{\mathbf{p}}_j$ by tree $t$ is the error for joint $j$.
Again we denote as $\mathcal{T}_d$ the forest created by this discriminative training.

\subsubsection*{Regression forest: Leaf Nodes}
\label{sec:reg_leaf_node}

Denote as $S$ the set of training examples arriving at leaf node $l$,
let $i \in S_j \subseteq S$ indexes over the subset $S_j$ containing training examples that are sufficiently close to joint $j$ in 3D space,
and let $\tilde{i} \in S \setminus S_j$ indexes over the complementary subset.
Let $\mathbf{o}_{li \rightarrow j}$ represent the offset of a particular training example $i$ of leaf node $l$ to joint $j$.

For each joint $j$ of the mouse visible from top-mounted camera, the mean $\mu_j$ of the offset vectors $\mathbf{o}_{li \rightarrow j}$ is computed and stored in the leaf.
There might be few cases in training when the offsets are distributed too widely or are multi-modal and might not be useful as predictor in leaf.
One solution to this problem is to compute the meanshift of the offset vectors.
Another alternative is to use the largest eigenvalue $\lambda_1$.
When $\lambda_1 < \lambda_{max}$, where $\lambda_{max}$ is a bound on the eigenvalue, we mark the offset vector stored in the leaf as a low confidence estimate.
At runtime, these estimates are used only if no other high confidence estimates available for any sampled pixel from the depth image from any other tree in the forest.

At test run every depth pixel sampled from the mouse patch will be processed through each of the trees in $\mathcal{T}_d$ from root to a certain leaf node $l$.
For each joint $j$ visible from top camera, the offset estimations from the leaf nodes are collected and their mean is used as the estimated offset position of joint $j$ from depth pixel location.

\subsubsection*{Discriminative training of $\mathcal{T}$}
\label{sec:reg_dis_train}

The discriminative training approach introduced in Sec.~\ref{sec:dis_train} can be adapted to this regression forest with a few modifications.
In practice, for $S \subseteq D'_d$ that arrives at a node $q$ having test $(\phi_q, \gamma_q)$, a set of $m$ tests $(\phi, \gamma)$ is attempted to maximize $\mathcal{E}_{(\phi, \gamma)}(t, D'_d)$.
Our discriminative training method changes the nodes iteratively in this manner while attempting to maintain the existing structure of the tree if that gives better performance.
There are two exceptional cases as before.
If $|S| \leq l_n$ at a split node $q$, then the subtree rooted at $q$ is replaced with a leaf node that maximizes $\mathcal{E}$.
In the second case, if $|S| > l_n$ at a leaf node $q$, then a subtree is dynamically grown and it is replaced in the place of $q$.
We denote as $\mathcal{T}_d$ the forest created by this discriminative training.

\subsection*{Step 2: Paw Detection from Bottom-view Color Image}


From top-mounted depth camera, we are now able to estimate the main body joints 1-12, as displayed in Fig.\ref{fig:anatomy-b}.
However, this is insufficient for full-body joint estimation since the limbs and paws are typically occluded below the main body.
This issue would persist with alternate setups such as a side-mounted depth camera.
This inspires us to place an additional color camera below the cage to explicitly aid the estimation of lower limbs and paws.

Our paw detector utilizes a cascade Adaboost classifier with Haar wavelet features, similar to~\cite{VioJon:ijcv04}.
Our cascade contains 20 stages that have been trained using mice images captured from bottom camera, with regions of paws being manually cropped and used as foreground bounding boxes.
Background images are obtained by randomly cropping from the static background images with diverse lighting conditions and the paw regions blurred out.
Empirical evidence discussed in later section also demonstrates the effectiveness of the proposed paw detector in our context.

\subsection*{Step 3: Fusion by 3D Calibration and Registration}
\label{sec:unification}

Step 3 integrates the intermediate outputs of Steps 1 and 2 to deliver the full body pose of mouse.
A global 3D coordinate system and associated transformations are required to align the intermediate outputs.
Chessboard is used to calibrate the intrinsic parameters of the two camera separately.
Extrinsic parameters are computed by solving the Perspective-n-Point (PnP) problem~\cite{LiXuXie:pami12}.
These parameters are used to transform the 1-12 joints estimated from top-view and the paw joints 21-24 from the bottom view to the same global coordinate system.

The paws are assumed to be on or close to the floor.
This is usually true, except in grooming or rearing poses, where the front paws can be observed from the top view.
If the front paws cannot be observed from either camera, a default configuration of the missing joints will be imposed by executing the kinematic chain model.

\begin{figure}[!tp]
\centering
\includegraphics[scale=.6]{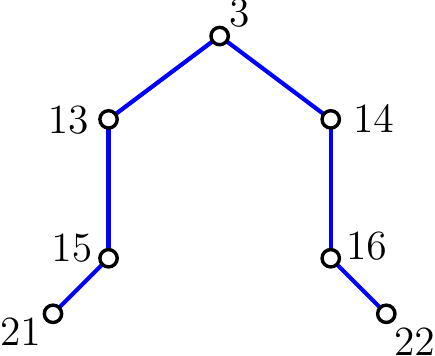}
\hspace{5mm}
\includegraphics[scale=.6]{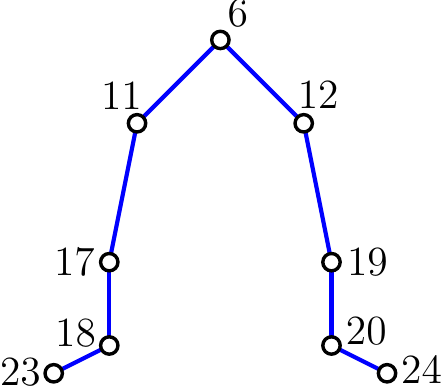}
\caption
{
    \footnotesize{Left: Fore limb structure. Right: Hind limb structure.}
}
\label{fig:limb}
\end{figure}

What remains to be computed are the limb joints that connect the main body to the paws.
The forelimb joints 13-16, seen in Fig.\ref{fig:limb}(a), can be estimated by local inverse kinematics, using the paw locations and the lengths of limb bones.
The left hind joints, seen in Fig.\ref{fig:limb}(b), are solved by setting the angle between 17, 18, and 23 to $90\degree$, and computing the location of 17 using local inverse kinematics.
The right hind joints are computed similarly.

\section{Experiments}

The proposed pose estimation approach has been applied to a variety of synthetic and real data, to different types of rodents, to different types of depth cameras and imaging setups.
In addition, to demonstrate the applicability of our approach to work on related problems, our method is used for the task of part-based labeling from single depth images, where it is shown to produce satisfactory results.

The training and testing are performed on a computer with Intel Core i7-4790 CPU and 16 GB of memory.
Our discriminative training stage is multi-threaded and the rest of the steps use single CPU core.
The full-body pose estimation from a pair of depth and color images on real data is currently unoptimized and performs at a speed of 2 frame-per-second (FPS).

Unless mentioned otherwise, all the regression forests are trained with $|\mathcal{T}|=7$ and $|\mathcal{T}_d|=7$.
A set of $240,000$ synthetic images generated using common mouse poses together with pitch, roll, and in-plane rotations and scale variations of the mouse model is used as training data.
The depth pixels for training, discriminative training and testing are randomly sampled from these images.
At each split node, a subset of $m = 50$ tests are randomly selected.
$\epsilon = 10$ is used for paws, $\epsilon = 15$ for limb joints and ears, $\epsilon = 50$ for tail joints and $\epsilon = 25$ for the rest.
$\tau = 10$ is used for weights at leaf nodes.
Trees are grown to a maximum of $L = 20$ levels or a node has less than $l_n = 60$ training examples.
For paw detection, a cascaded classifier with 20 stages is trained using 4200 images with paws manually marked and 3000 negative background images.

\subsection{Synthetic datasets for mouse pose estimation}\label{sec:synthetic}

To evaluate our method we use joint error as defined in Sec.~\ref{sec:dis_train}.
We scale the model and the joint error assuming a mouse length of $100$mm, which is about the length of an adult mouse.
As test data, 700 images are generated with variations in mouse size, pose, and orientation.

\subsubsection{Effect of forest size}

\begin{figure}[!tp]
\centering
\includegraphics[width=0.9\linewidth]{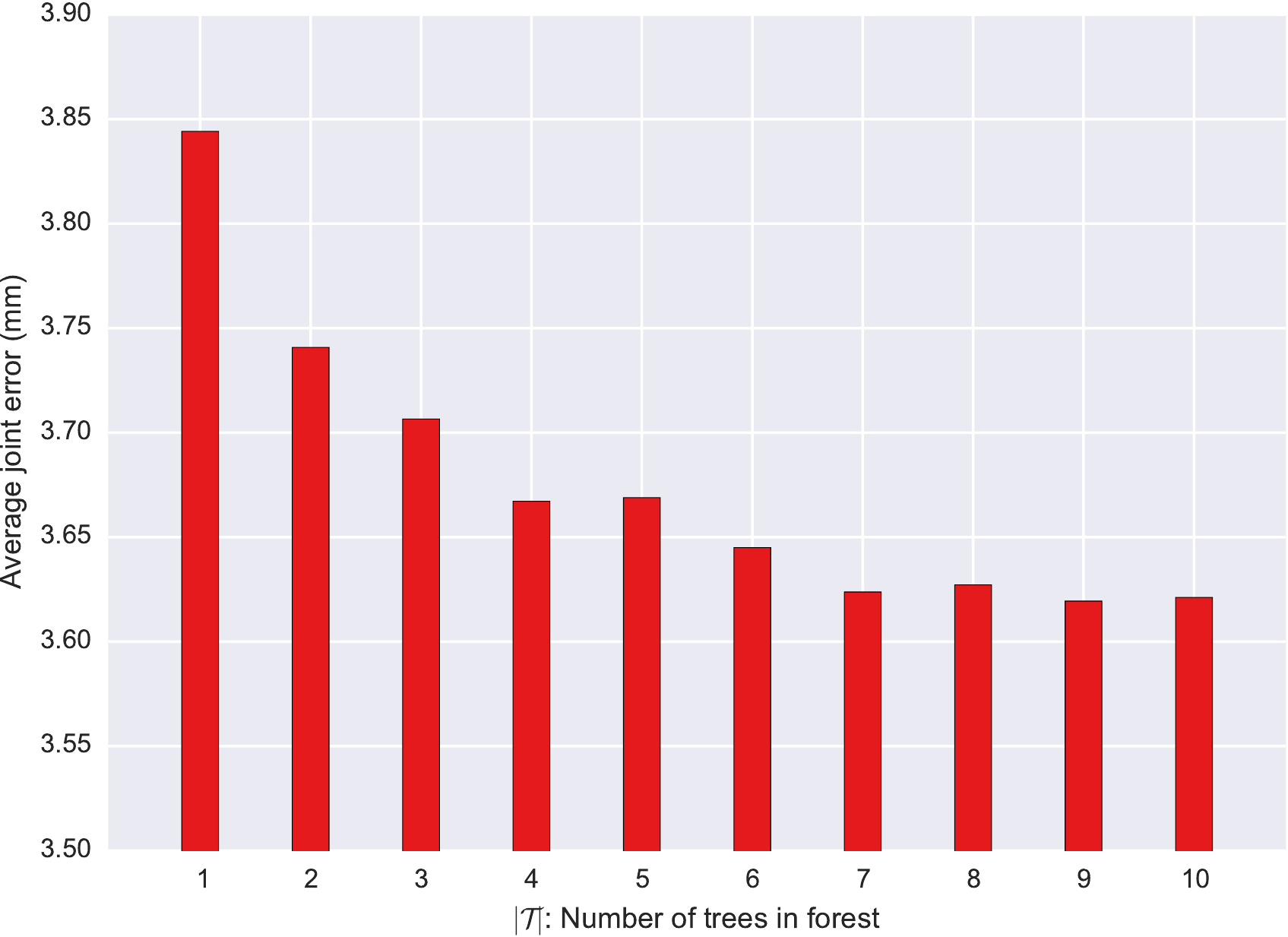}
\caption{\footnotesize{Effect of forest size in $\mathcal{T}$ on average joint error.}}
\label{fig:reg_tree_num}
\end{figure}

In Fig.~\ref{fig:reg_tree_num} we look at the effect of forest size in $\mathcal{T}$ on the average joint error for the $12$ joints estimated from top depth image.
The values presented are the average difference in Euclidean distance between estimated and true joint positions in 3D over all joints.
We can see that the joint error decreases as the number of trees in $\mathcal{T}$ increases.
However, the joint error does not decrease much after $|\mathcal{T}| = 7$, where the average joint error is $3.62$mm.
We will use this forest size for the rest of our experiments with pose estimation.

\subsubsection{Effect of $m$}

\begin{figure}[!tp]
\centering
\includegraphics[width=0.9\linewidth]{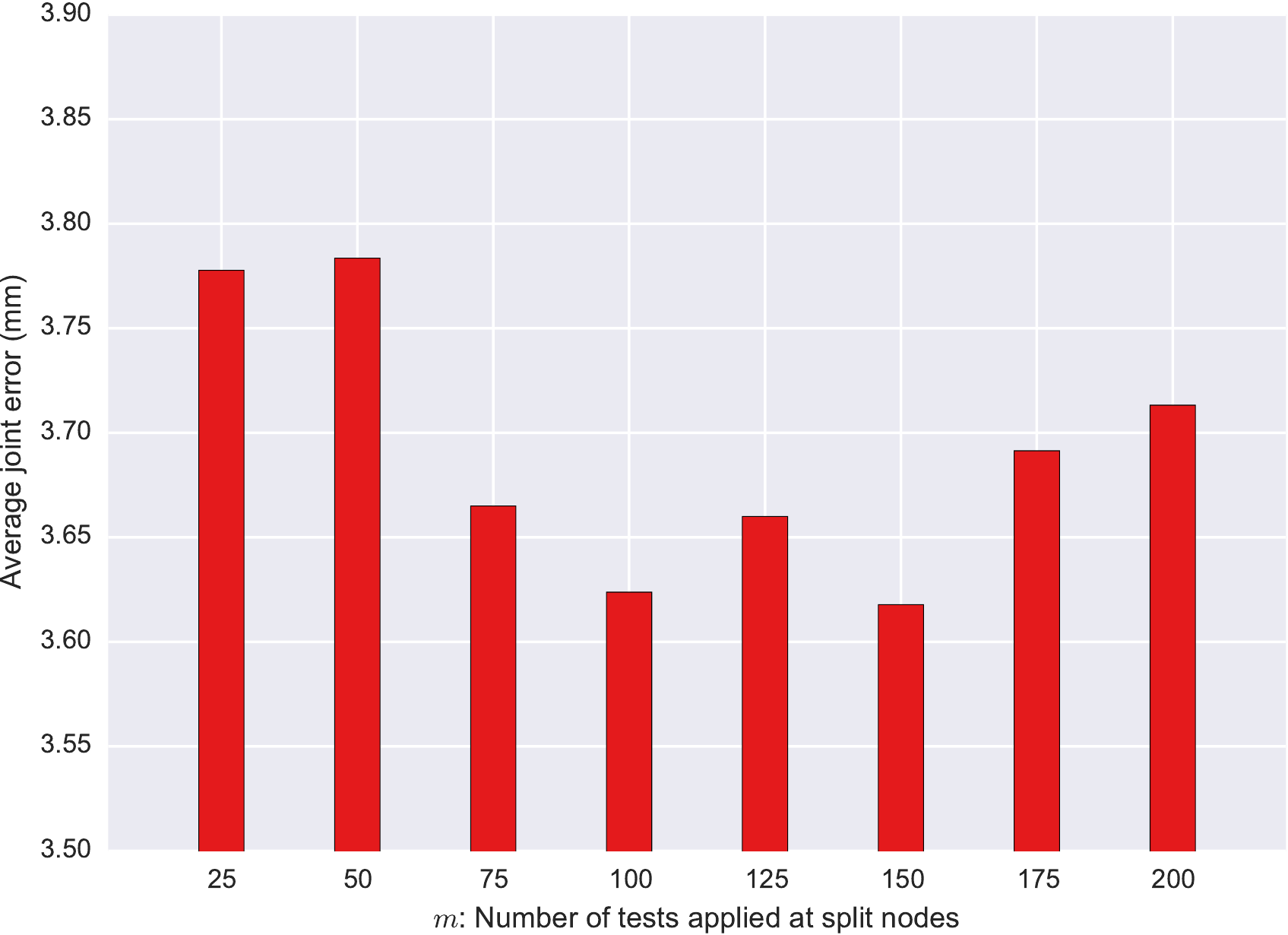}
\caption{\footnotesize{Effect of $m$ in $\mathcal{T}_d$ on average joint error.}}
\label{fig:reg_test_num}
\end{figure}

In Fig.~\ref{fig:reg_test_num} we examine the effect of $m$, the number of tests applied at split nodes in $\mathcal{T}$. 
We see that the average joint error over all estimated joints decreases with increasing values of $m$ until $m = 100$, after which it increases.
This can be attributed to the increasing correlation between the split nodes of different trees in the forest as the sampling size from a fixed number of tests increases.
We use $m = 100$ for the rest of our experiments with pose estimation.

\subsubsection{Effect of discriminative training}

\begin{figure}[!tp]
\centering
\includegraphics[width=0.9\linewidth]{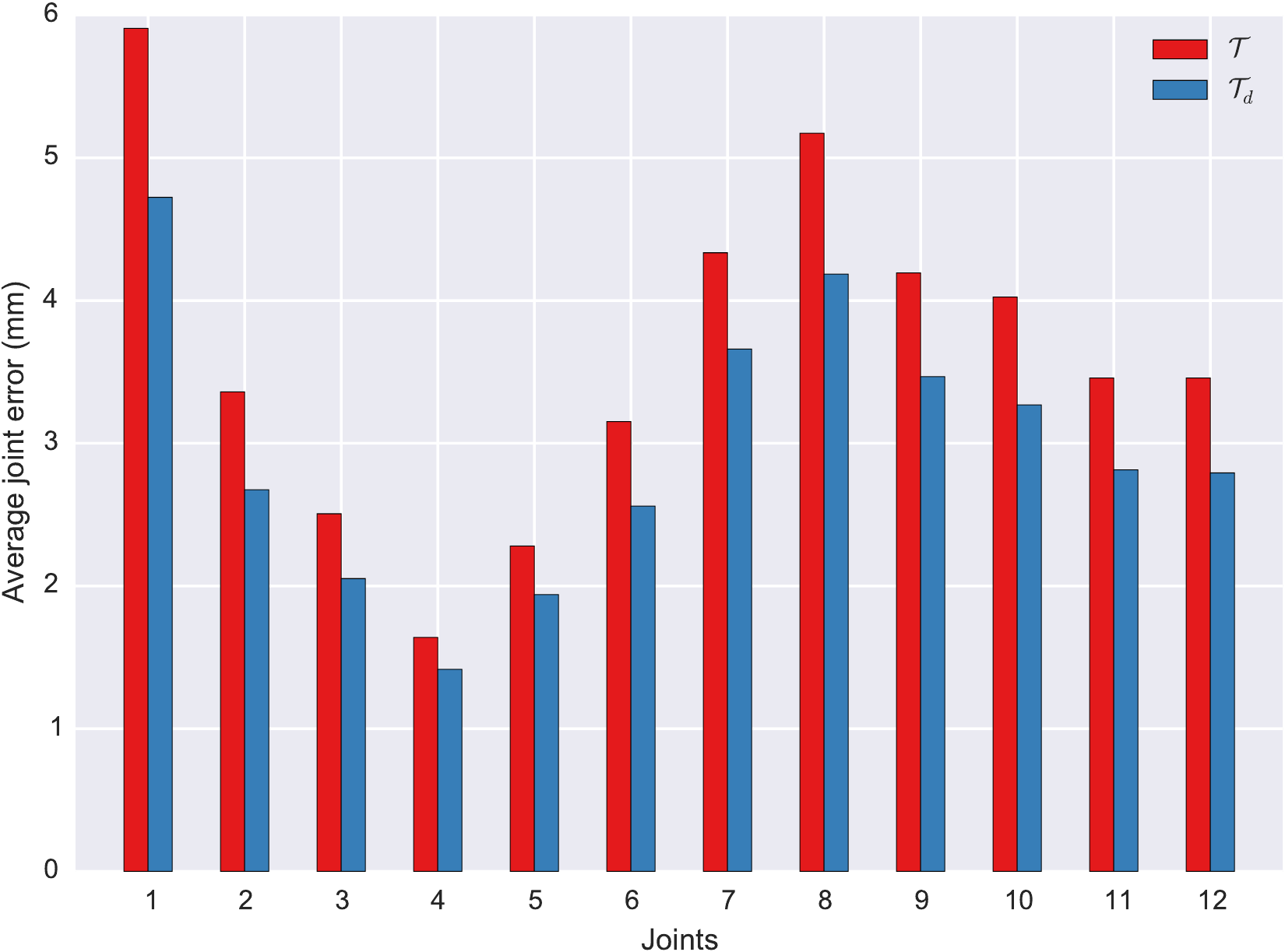}
\caption{\footnotesize{Comparison of joint error between baseline $\mathcal{T}$ and $\mathcal{T}_d$ trees.}}
\label{fig:reg_dis}
\end{figure}

In Fig.~\ref{fig:reg_dis} we examine the effect of applying discriminative training as described in Sec.~\ref{sec:reg_split_node} with naturally the same number of trees $|\mathcal{T}| = |\mathcal{T}_d| = 7$ and $m = 100$.
The graph shows a comparison of the errors for the $12$ joints estimated from top camera synthetic depth image for both $\mathcal{T}$ and $\mathcal{T}_d$.
We can see that this discriminative learning process has substantially decreased the error on every joint.
For example, joint $1$ which has the highest error of $5.90$mm with $\mathcal{T}$ is reduced to $4.72$mm with $\mathcal{T}_d$.
The average error over all joints for $\mathcal{T}$ is $3.62$mm and it is substantially reduced to $2.96$mm by $\mathcal{T}_d$.
For all the following experiments we use this discriminatively trained $\mathcal{T}_d$.

%
%

\begin{figure}[!tp]
\centering
\includegraphics[width=0.9\linewidth]{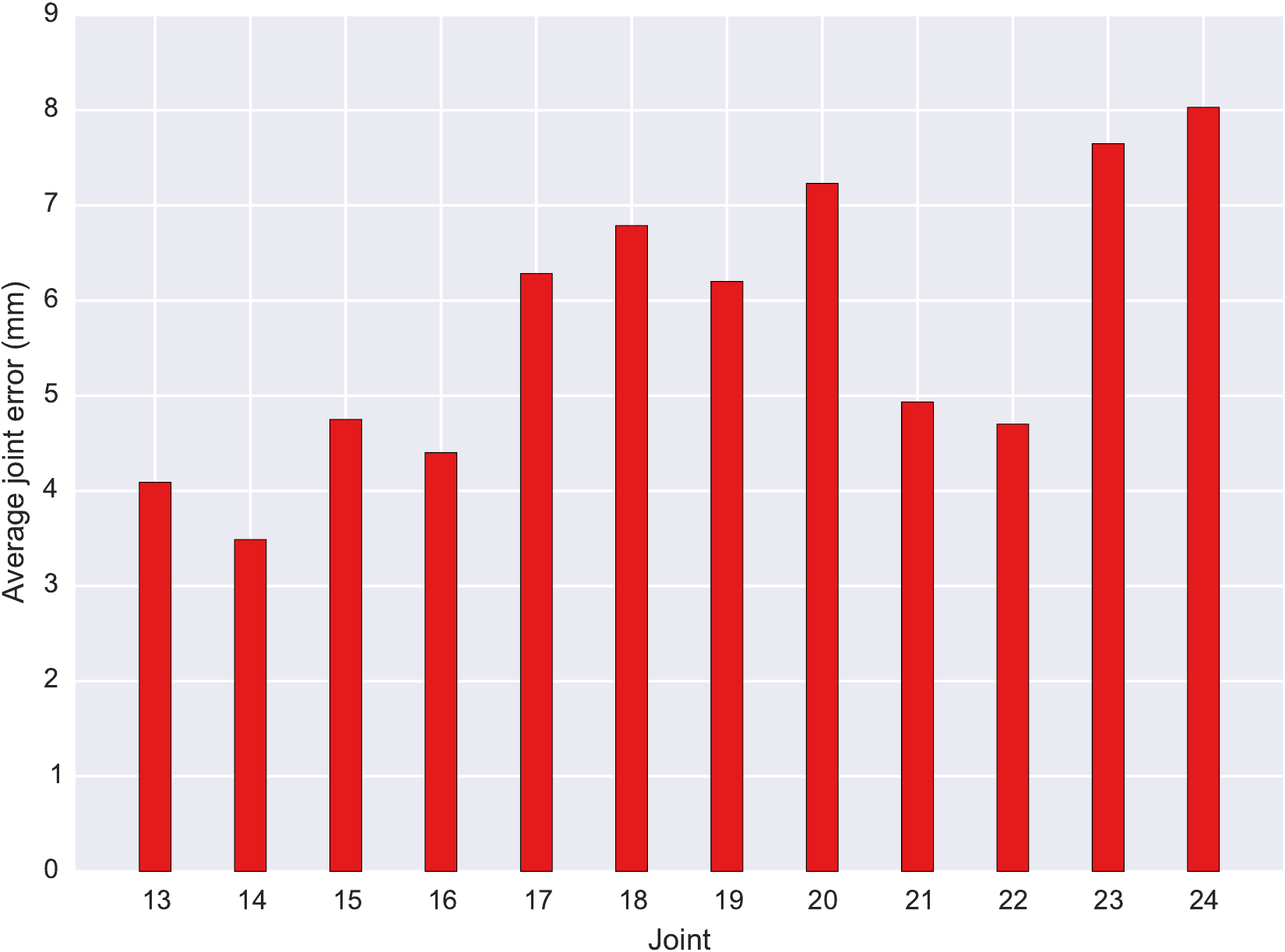}
\caption{\footnotesize{Average error of joint positions estimated from bottom image.}}
\label{fig:joint_error}
\end{figure}

Fig.~\ref{fig:joint_error} shows the average joint error for the limb and paw joints estimated using the paw locations detected in bottom image.
It can be observed that these errors are relatively higher compared to top joints.
This is due to the higher DoF of limb joints compared to spine and our method of limb joint estimation is only a relatively coarse approximate.

\begin{figure*}[!tp]
\centering
\subfigure[]
{
    \includegraphics[width=0.97\linewidth]{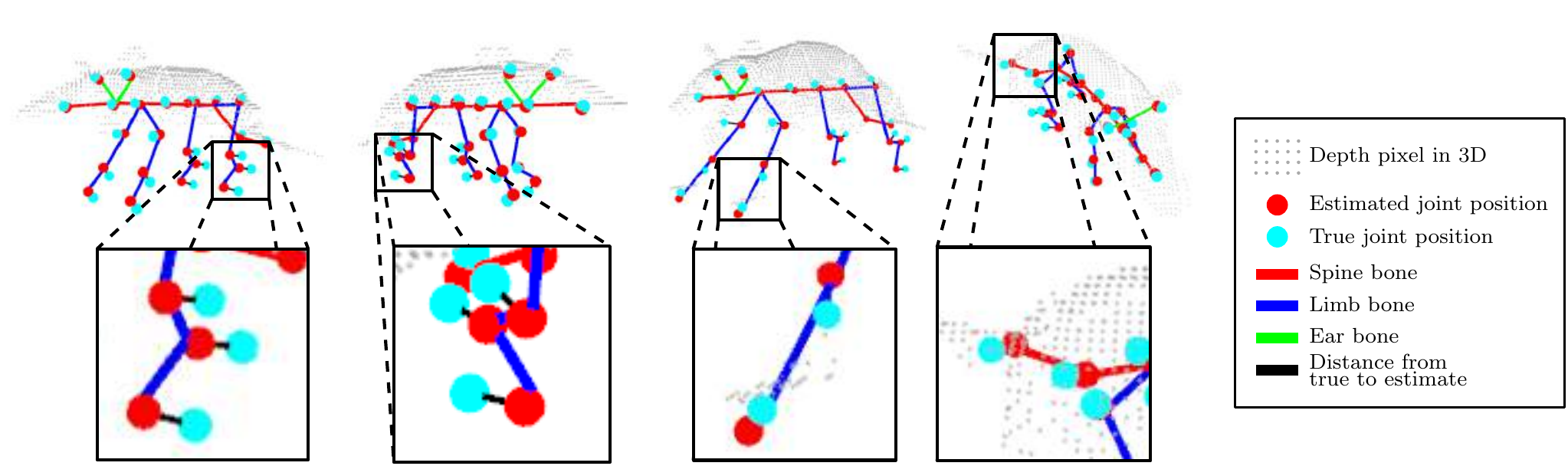}
}
\hspace{5mm}
\subfigure[]
{
    \includegraphics[width=0.95\linewidth]{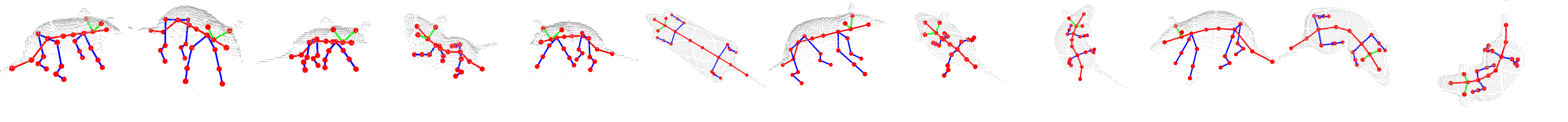}
}
\caption
{
    \footnotesize
    {
        Top: Four pose estimates are obtained by applying our method on depth-color test-image pairs containing poses of standing, walking, running and bending while walking respectively.
        The estimated and true joint positions are shown as \textcolor{red}{red} and \textcolor{cyan}{cyan} disks respectively.
        The depth pixels converted to 3D positions are rendered as grey dots.
        Bones of the skeleton are colored following the protocol of Fig.\ref{fig:anatomy-b}.
        Bottom: A diverse range of estimated poses as seen from various viewpoints in 3D.
    }
}
\label{fig:synthetic}
\end{figure*}

Fig.\ref{fig:synthetic} presents several results of utilizing $\mathcal{T}_d$ for mouse pose estimation with diverse gestures and orientations.
The figure shows the actual 3D positions of both the ground-truth and estimated joints.
Visually our estimation results are shown to be well-aligned with the ground-truth poses.

\begin{figure*}[!tp]
\centering
\includegraphics[width=0.99\linewidth]{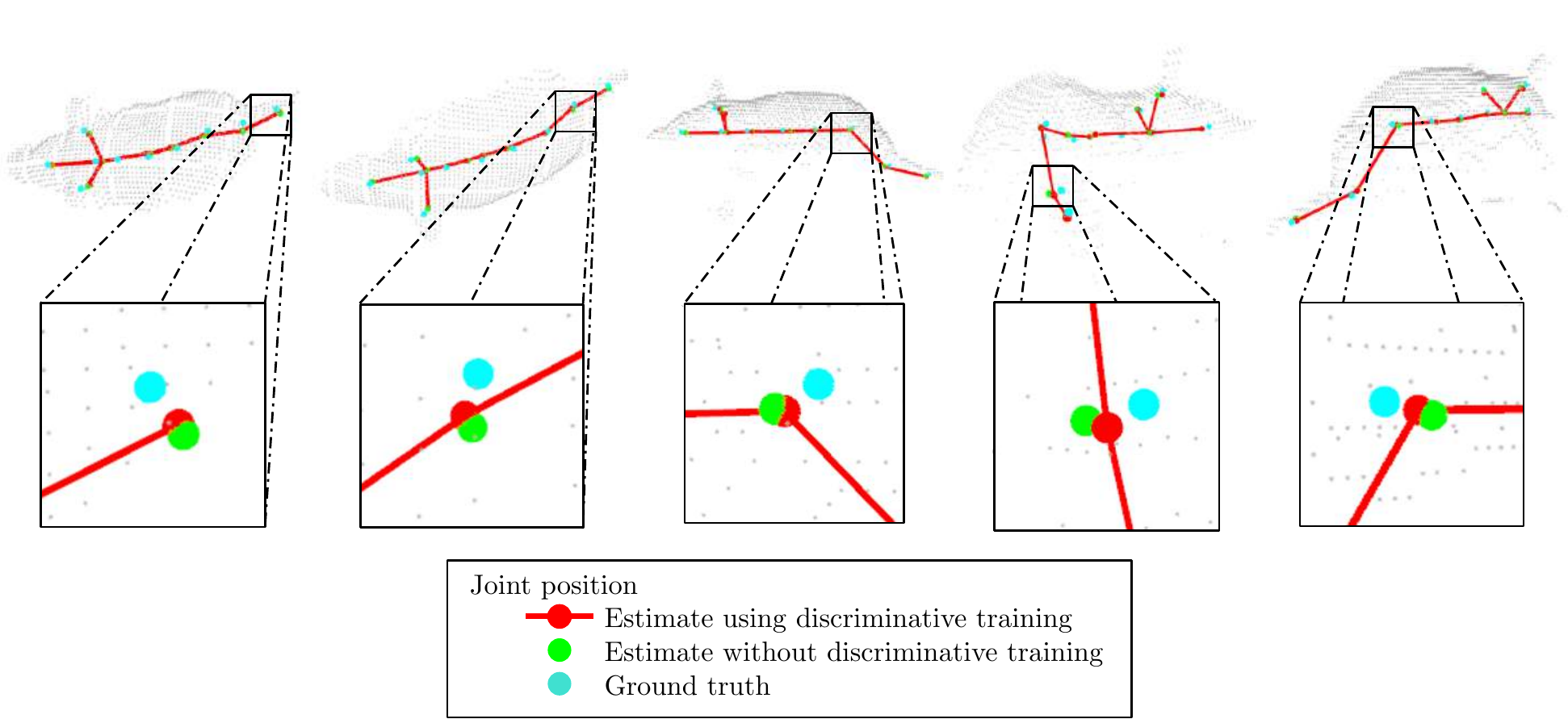}
\caption
{
    \footnotesize
    {
        Comparison of pose estimation results using forests without discriminative training and with discriminative training.
        Results show that with discriminative training, the estimated joint positions are closer to the ground-truth and have lesser error than without discriminative training.
    }
}
\label{fig:dis_train_pose}
\end{figure*}

Fig.~\ref{fig:dis_train_pose} shows a few exemplar results comparing the pose estimation results on synthetic data using forests without and with discriminative training.
The joint positions, drawn as spheres, are colored based on whether they are ground-truth, estimated with and without discriminative training.
It can be observed that discriminative training helps in reducing the joint position error and the estimated joints lie closer to ground-truth.

\subsubsection{Effect of signal noise}

\begin{figure}[!tp]
\centering
\includegraphics[width=.4\textwidth]{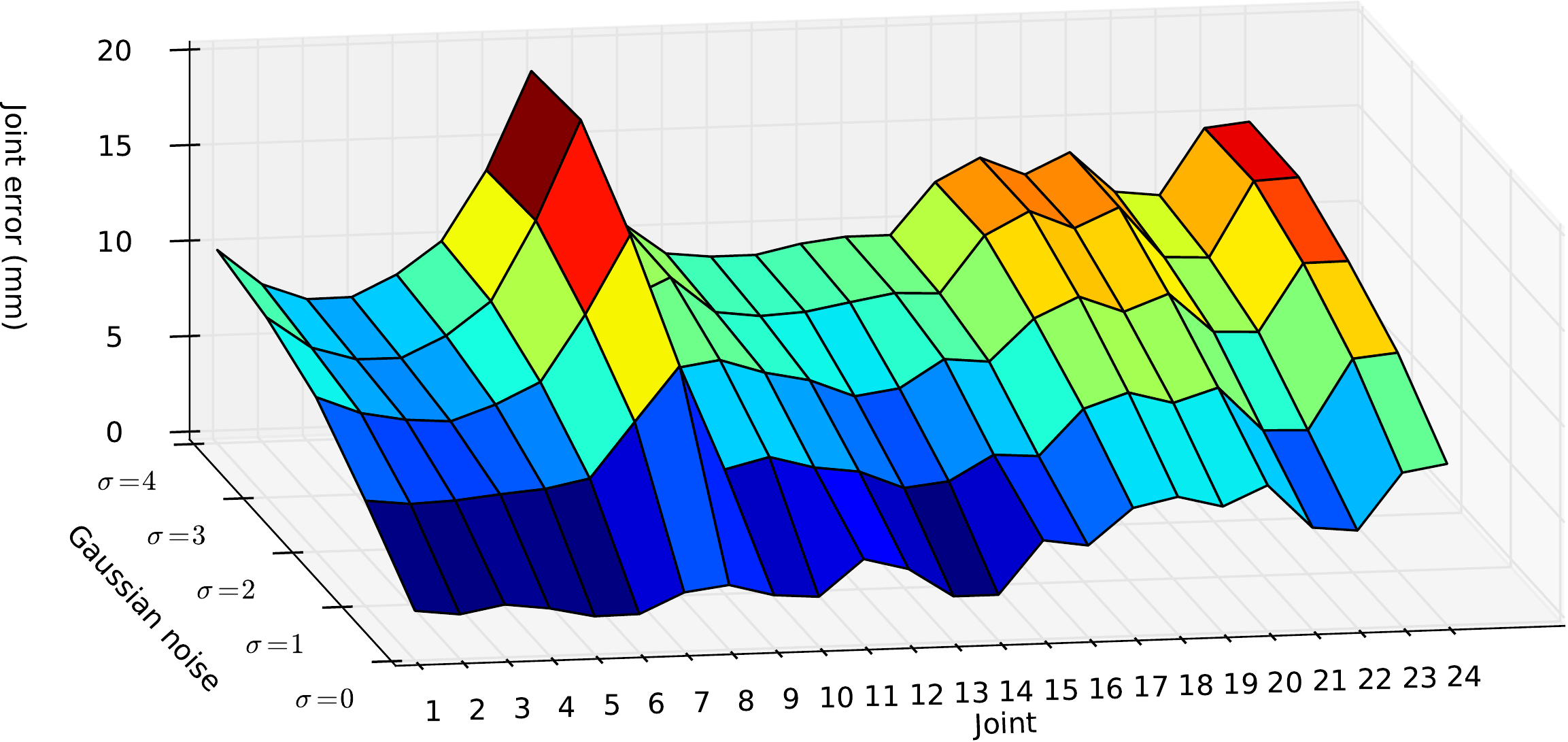}
\caption
{
    \footnotesize
    {
        Average joint error with increasing amount of Gaussian noise $\sigma = \{0, 1, 2, 3, 4, 5\}$ added to input depth images.
    }
}
\label{fig:joint_error_noise_mshift}
\end{figure}

Fig.\ref{fig:joint_error_noise_mshift} shows the robustness of our method to noise in depth image.
Gaussian random noise with levels $\sigma = \{0, 1, 2, 3, 4, 5\}$ is applied to synthetic depth images.
We see that there is only a slight increase in average joint error in proportion to the noise.
This demonstrates the robustness of our random forest model for joint position estimation.
The tail joint is highly sensitive to noise, which can be attributed to its appearance as a thin strip of pixels in the depth image and its topology can be warped easily by noise.

%

\subsection{Real datasets for mouse pose estimation}

Real data have been captured with two rodents: lab mouse (\textit{mus musculus}) and hamster (\textit{cricetidae}).
Two consumer depth cameras are employed for top-view: structured illumination (Primesense Carmine) and ToF (SoftKinetic DS325).
Though the noise characteristics of the two depth cameras are significantly different, our approach utilizing $\mathcal{T}_d$ is able to reliably deliver good results.
The color images are all from a bottom-view Basler CMOS camera. 

\subsubsection{Pose estimation from a pair of depth and color images}

\begin{figure*}[!tp]
\centering
\includegraphics[width=0.9\linewidth]{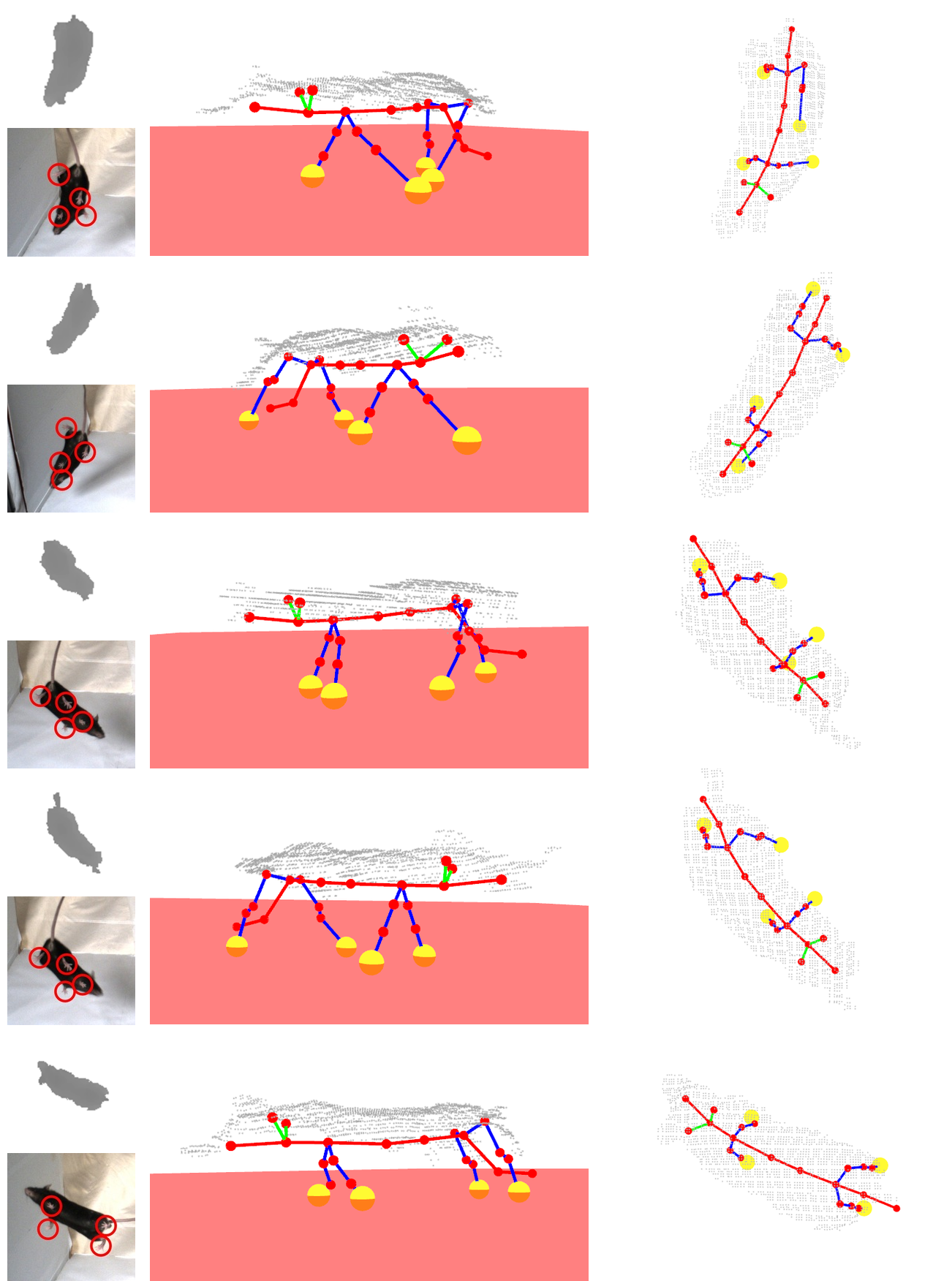}
\caption
{
    \footnotesize
    {
    Exemplar results of full-body pose estimation of lab mouse (\textit{mus musculus}) from image pairs captured from Primesense depth and Basler color cameras.
    Left column shows pairs of cropped regions from depth and color input images with the detected paws marked in \textcolor{red}{red} circles.
    Middle column shows a side-view 3D rendering of the estimated mouse pose skeleton while right column is the corresponding top view rendering of the pose.
    Bones of the skeleton are colored following the protocol of Fig.~\ref{fig:anatomy-b} in our paper.
    The paws are rendered as yellow spheres. The depth pixels (i.e. 3D cloud of points) are rendered in light-grey and floor is rendered in pink color.
    }
}
\label{fig:real_results_lab_mouse}
\end{figure*}

\begin{figure*}[!tp]
\centering
\includegraphics[width=0.99\linewidth]{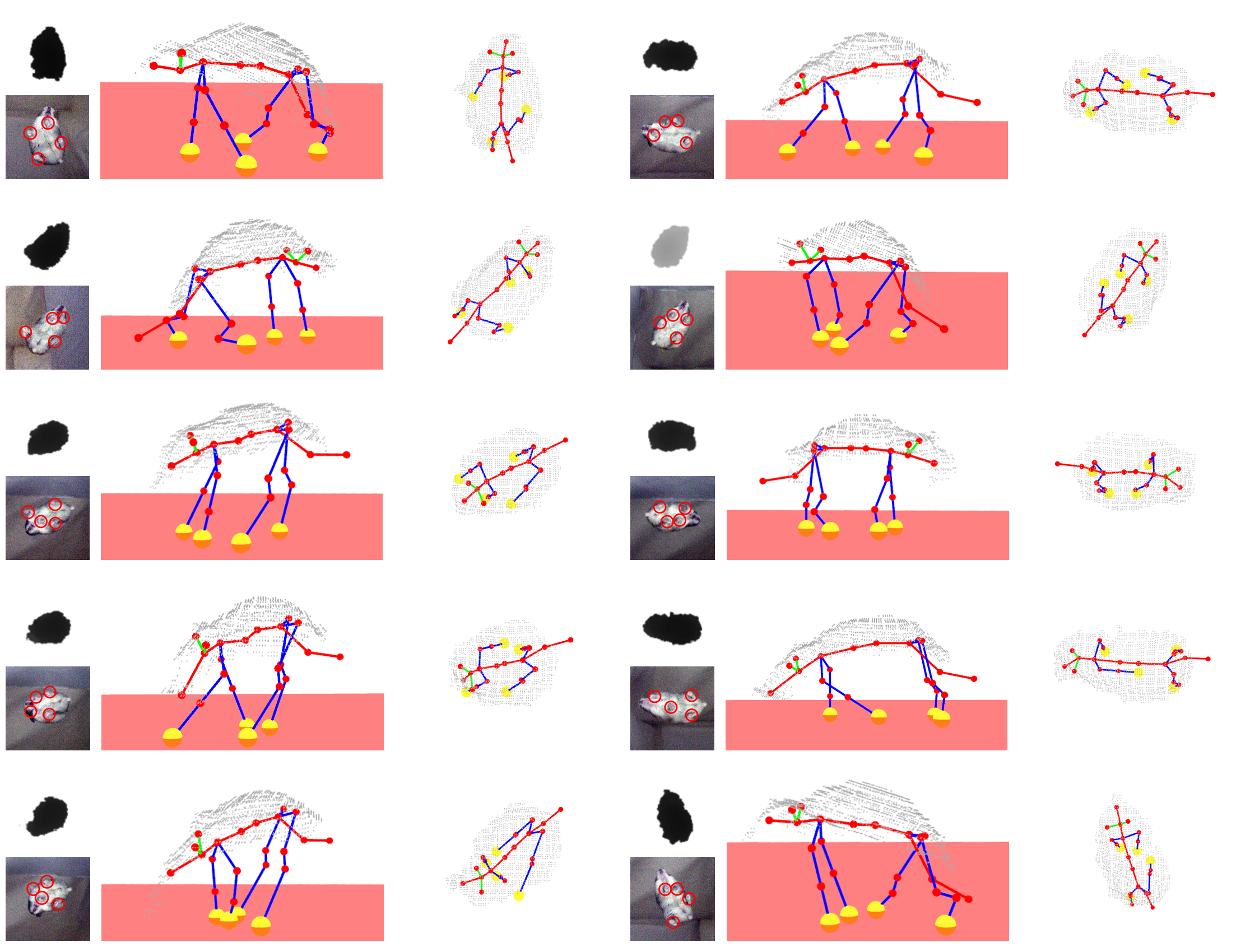}
\caption
{
    \footnotesize
    {
    Exemplar results of full-body pose estimation of hamster (\textit{cricetidae}) from image pairs captured from Primesense depth and Basler color cameras.
    Left column shows pairs of cropped regions from depth and color input images with the detected paws marked in \textcolor{red}{red} circles.
    Middle column shows a side-view 3D rendering of the estimated mouse pose skeleton while right column is the corresponding top view rendering of the pose.
    Bones of the skeleton are colored following the protocol of Fig.~\ref{fig:anatomy-b} in our paper.
    The paws are rendered as yellow spheres. The depth pixels (i.e. 3D cloud of points) are rendered in light-grey and floor is rendered in pink color.
    }
}
\label{fig:real_results_hamster}
\end{figure*}

Using the two-camera setup of Fig.~\ref{fig:pipeline}, a large number of image pairs are captured of both lab mouse and hamster.
Primesense depth camera is used for top-view and a Basler color camera is used for bottom-view.
Exemplar results on the image pairs of lab mouse (\textit{mus musculus}) is illustrated in Fig.\ref{fig:real_results_lab_mouse}.
Similarly, Fig.\ref{fig:real_results_hamster} shows exemplar results for our hamster data set.
These results show that our approach can deliver visually meaningful full body skeletal pose outputs that match well with the camera observations.
This estimation is far from being trivial, considering the difficult contexts where many intermediate joints such as limb joints 13-20 are usually occluded from both cameras, and the input signals are highly noisy.

\subsubsection{Pose estimation from single depth images}

\begin{figure*}[!tp]
\centering
\includegraphics[width=0.75\linewidth]{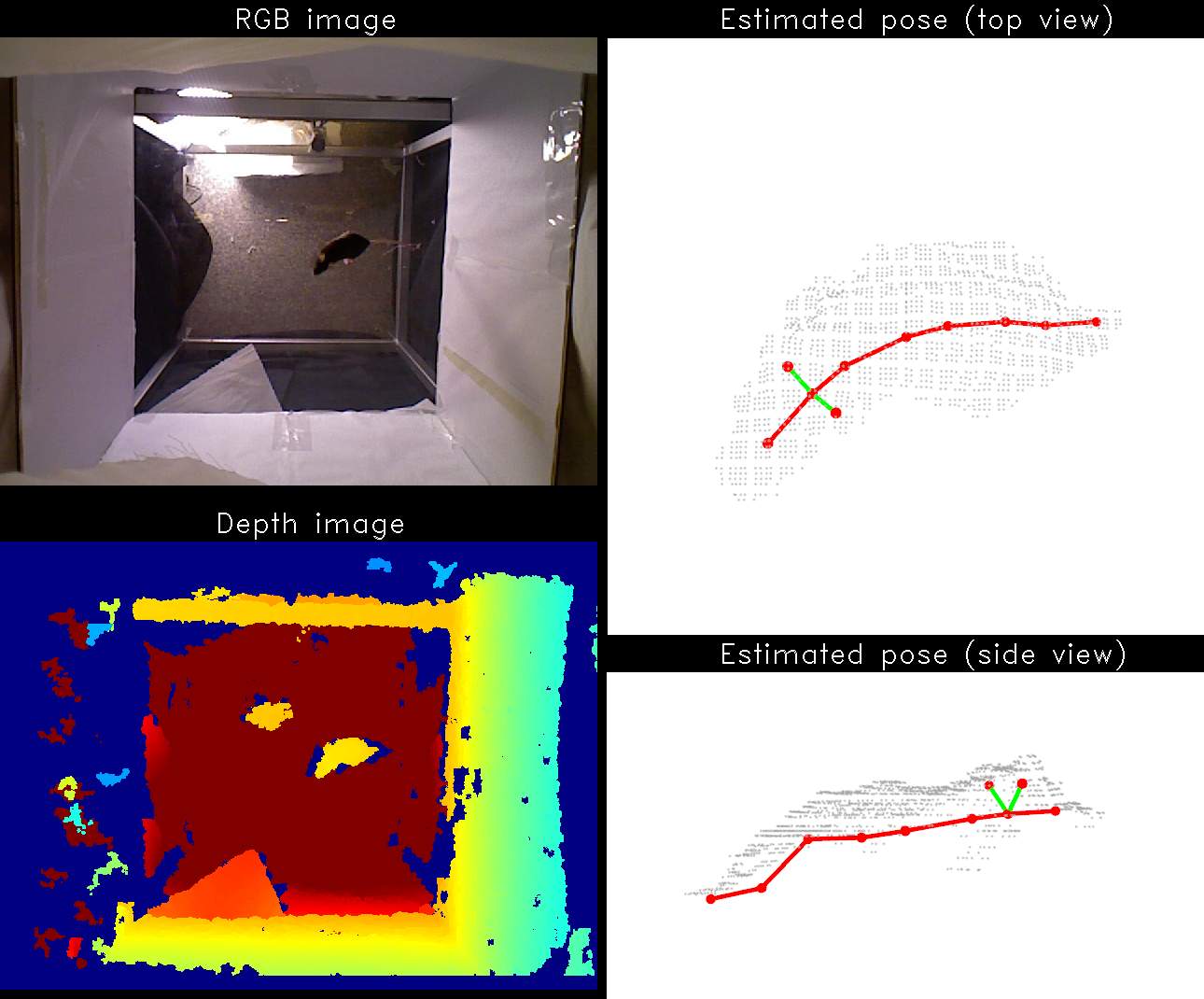}
\includegraphics[width=0.75\linewidth]{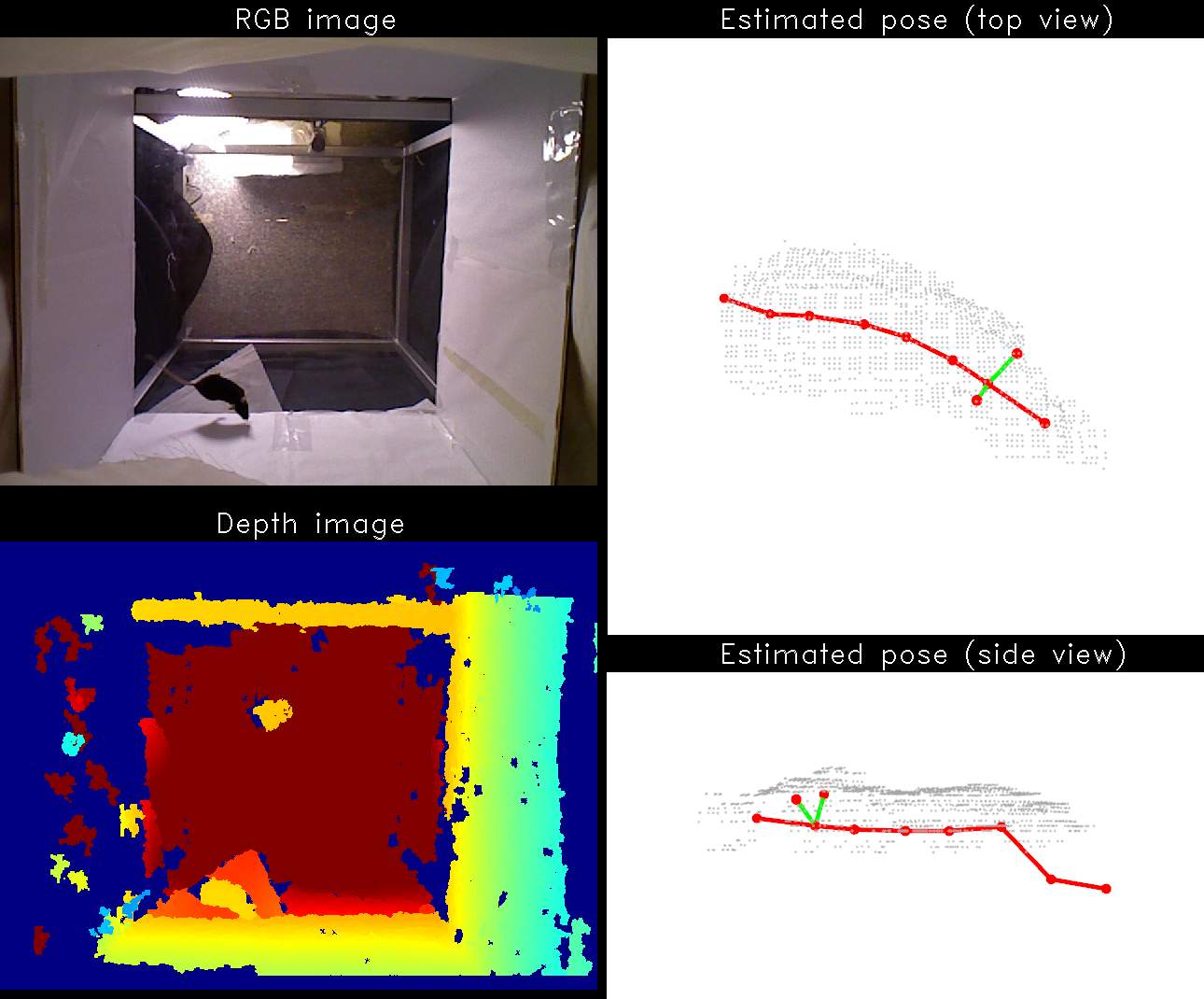}
\caption
{
    \footnotesize
    {
        Two examples of pose estimation based on single depth images from top-view Primesense structured illumination camera.
        The estimated 3D positions of spine joints are shown in top and side view.
        The white point cloud is the set of depth values of the mouse pixels obtained from the input depth image.
        Please see our supplementary video for more results on real data.
    }
}
\label{fig:single_primesense}
\end{figure*}

\begin{figure*}[!tp]
\centering
\includegraphics[width=0.99\linewidth]{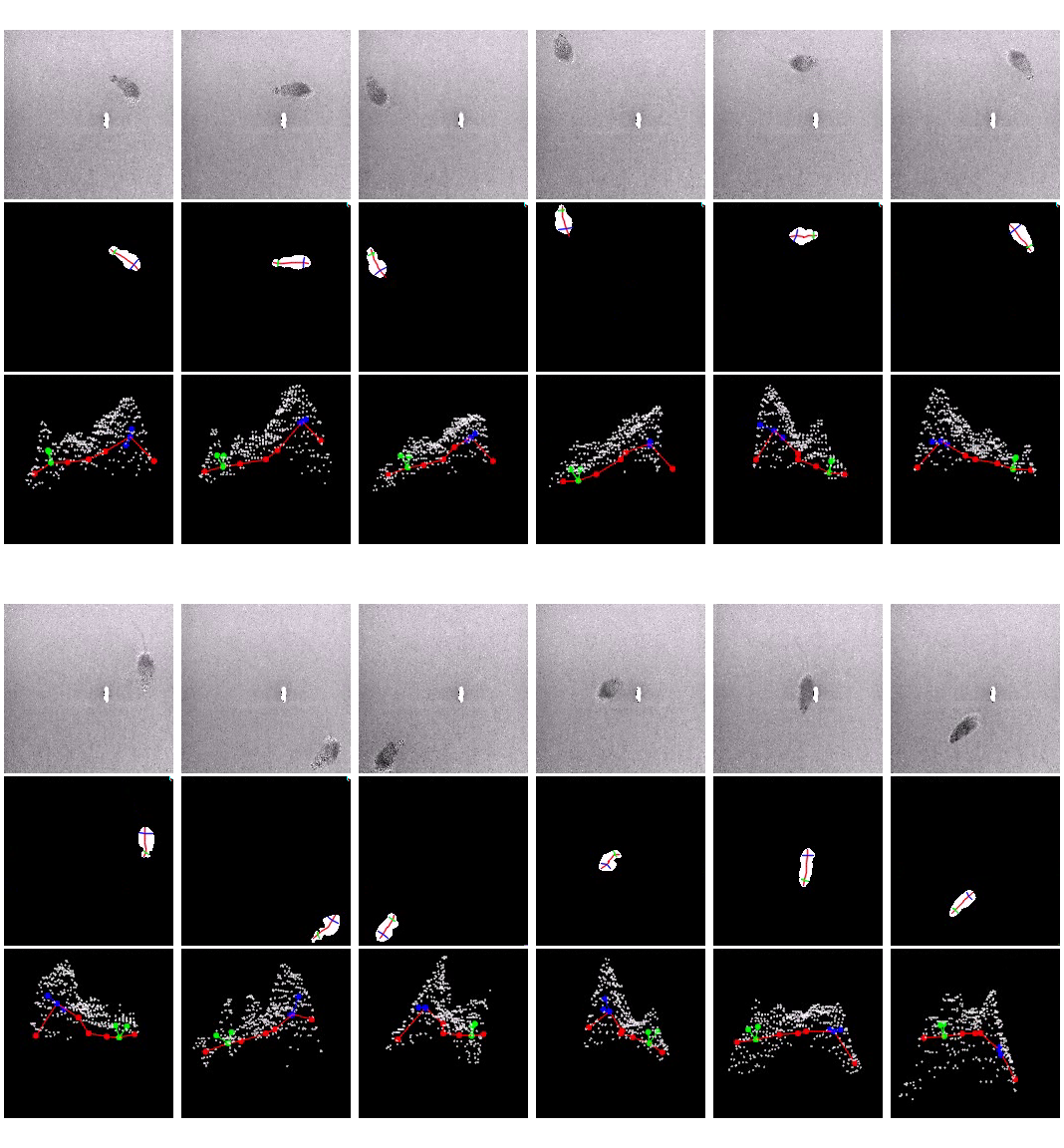}
\caption
{
    \footnotesize
    {
    Exemplar results of pose estimation based on single depth images from top-view SoftKinetic ToF camera.
    Top: Single depth images.
    Middle: Projection of the estimated joints’ 3D locations onto the top-view (i.e. the same viewing position as of the depth camera).
    Bottom: Projection onto a novel side view.
    The white point cloud is the set of depth values of the mouse pixels obtained from the input depth image.
    }
}
\label{fig:single_softkinetic}
\end{figure*}

To demonstrate the robustness of the proposed approach, a simpler setup of a single top-mounted depth camera is considered.
Using only the top-view depth images, the joints of the spine can be estimated by our random forest using only Step 1 of our system.

Fig.\ref{fig:single_primesense} shows our pose estimation results for top-view depth images obtained from a Primesense structured illumination camera.
In this case, only the estimated 3D positions of main body joints obtained from our discriminatively trained forest is used.

Fig.\ref{fig:single_softkinetic} shows our pose estimation results for top-view depth images obtained from a SoftKinetic DS325 ToF camera.
The resulting skeleton forms seem to be visually appealing, while remaining well-aligned with the inputs from both the input top-view as well as a novel side-view.
These results also demonstrate the general applicability of our system to depth cameras using different technologies like structured illumination and time-of-flight (ToF).

\begin{figure}[!tp]
\centering
\includegraphics[width=.9\linewidth]{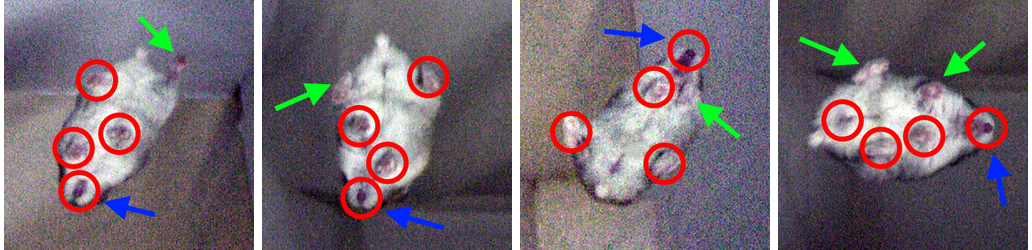}
\caption
{
    \footnotesize
    {
    Paw detection errors in depth images from Primesense camera.
    Detected paws are circled in \textcolor{red}{red}, wrongly detected region (usually mouth) is indicated with \textcolor{blue}{blue} arrow and undetected paw is shown with \textcolor{green}{green} arrow.
    }
\label{fig:real-bad}
}
\end{figure}

Fig.~\ref{fig:real-bad} shows examples of cases where paw detection has errors.
Sometimes the mouth or tail of the mouse is mistaken for a paw.

\subsection{Part-based labels from single depth images}
\label{sec:pbl_results}

Experiments are also carried out on the closely related task of part-based labeling of mouse using depth images captured from a top-mounted depth camera.
The body of the mouse is partitioned into six distinct classes with distinct colors as shown in Fig.\ref{fig:anatomy-d}.
A skin texture with these colors is used by our mouse engine to render poses to depth images.
During testing stage, every pixel of a test depth image is processed using the learned random forest model to obtain the averaged class histogram.
The estimated class for a depth pixel is the class with highest number of votes in the histogram.

\subsubsection{Synthetic datasets}

To generate training data, $360$ poses are rendered with random rotations applied on each pose.
A Gaussian noise of $\sigma = 16$ is applied on the depth images and a total of $496,800$ depth pixels are sampled.
A random forest $\mathcal{T}$ of 7 trees is trained using 20000 tests, $L = 13$ and $l_n = 60$.

To further improve the results of $\mathcal{T}$ on noisy depth images, we adapt the discriminative training approach described in Sec.~\ref{sec:dis_train} to perturb the nodes of $\mathcal{T}$ using a second training dataset $D_d$.
The nodes of these classification trees can be perturbed as described in Sec.~\ref{sec:reg_split_node} for regression trees, including grow and shrink actions as needed.
For classification, we try to maximize $\mathcal{E}(t, D_d)$ defined as:
\begin{align}
    \textstyle
    \mathcal{E}(t, D_d) = \sum_{i \in D_d} \mathcal{L}(\hat{y}_i, y_i),
\label{eq:Efunction}
\end{align}
where $\hat{y}_i$ is the predicted class for training example $i$ and $y_i$ is its true class.
$\mathcal{L} = 1$ when $\hat{y}_i = y_i$ and 0 otherwise.

\begin{figure*}[!tp]
\centering
\includegraphics[width=0.95\linewidth]{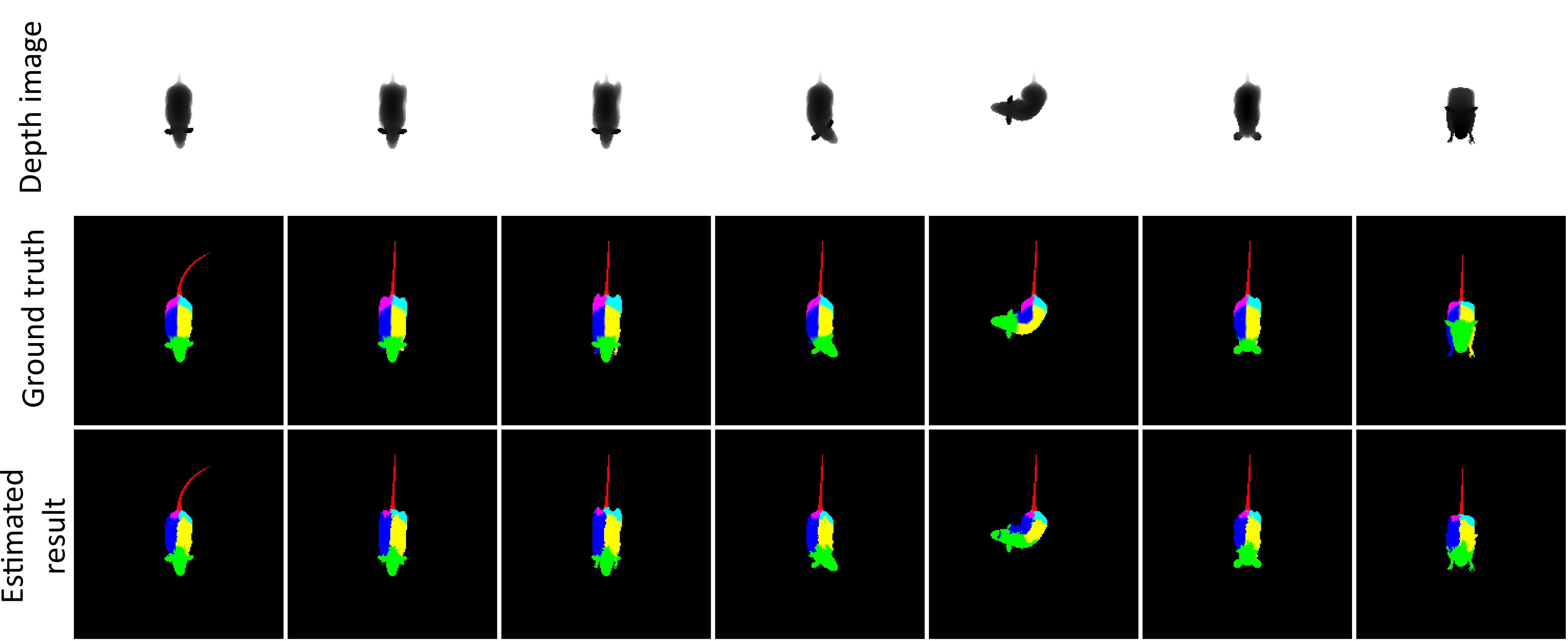}
\caption
{
    \footnotesize
    {
        Exemplar results of part-based labeling method. Top row is input depth image, middle row is ground truth and bottom row is estimated result from our discriminatively trained forest.
    }
}
\label{fig:pbl}
\end{figure*}

Fig.~\ref{fig:pbl} shows results of our part-based labeling method for a few exemplar inputs where the mouse is standing, walking, running, rearing and bending.
We can see that the estimated results match the ground truth for most of the pixels.

\begin{figure}[!tp]
\centering
\includegraphics[width=0.9\linewidth]{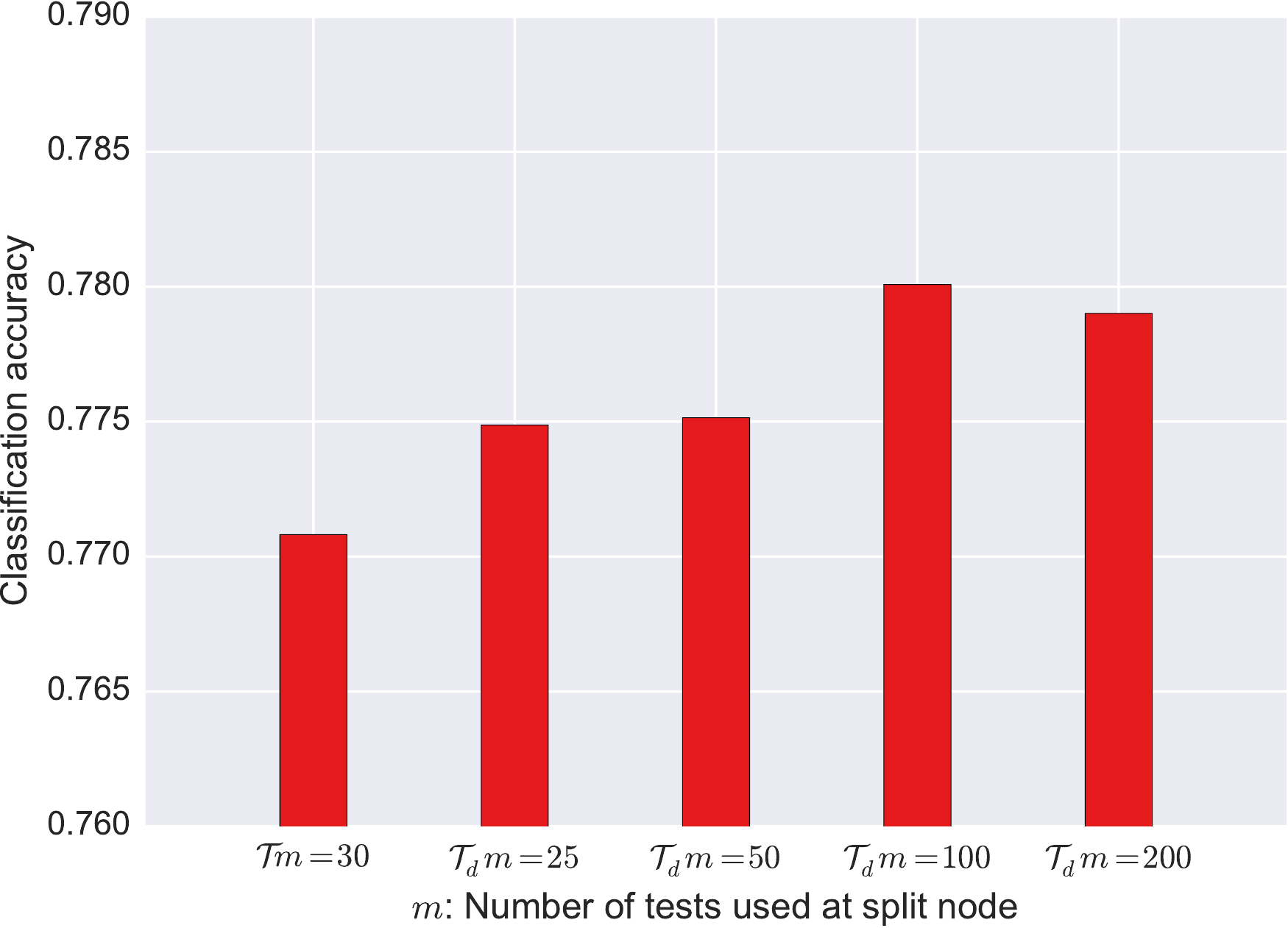}
\caption
{
    \footnotesize
    {
        Part-based labeling classification accuracy of base $\mathcal{T}$ with $m=30$ compared with discriminatively trained $\mathcal{T}_d$ with increasing values of $m$.
    }
}
\label{fig:pbl_perturb}
\end{figure}

Fig.~\ref{fig:pbl_perturb} shows the classification accuracy results for the forest before and after discriminative training of its nodes.
First we note that applying discriminative training gives better accuracy no matter what value of $m$ is used.
Further we notice that increasing $m$ increases accuracy, but only until $m = 100$ after which the accuracy reduces.
This can be attributed to the increased correlation between trees which decreases the performance of the forest.

\begin{figure*}[!tp]
\centering
\includegraphics[width=0.75\linewidth]{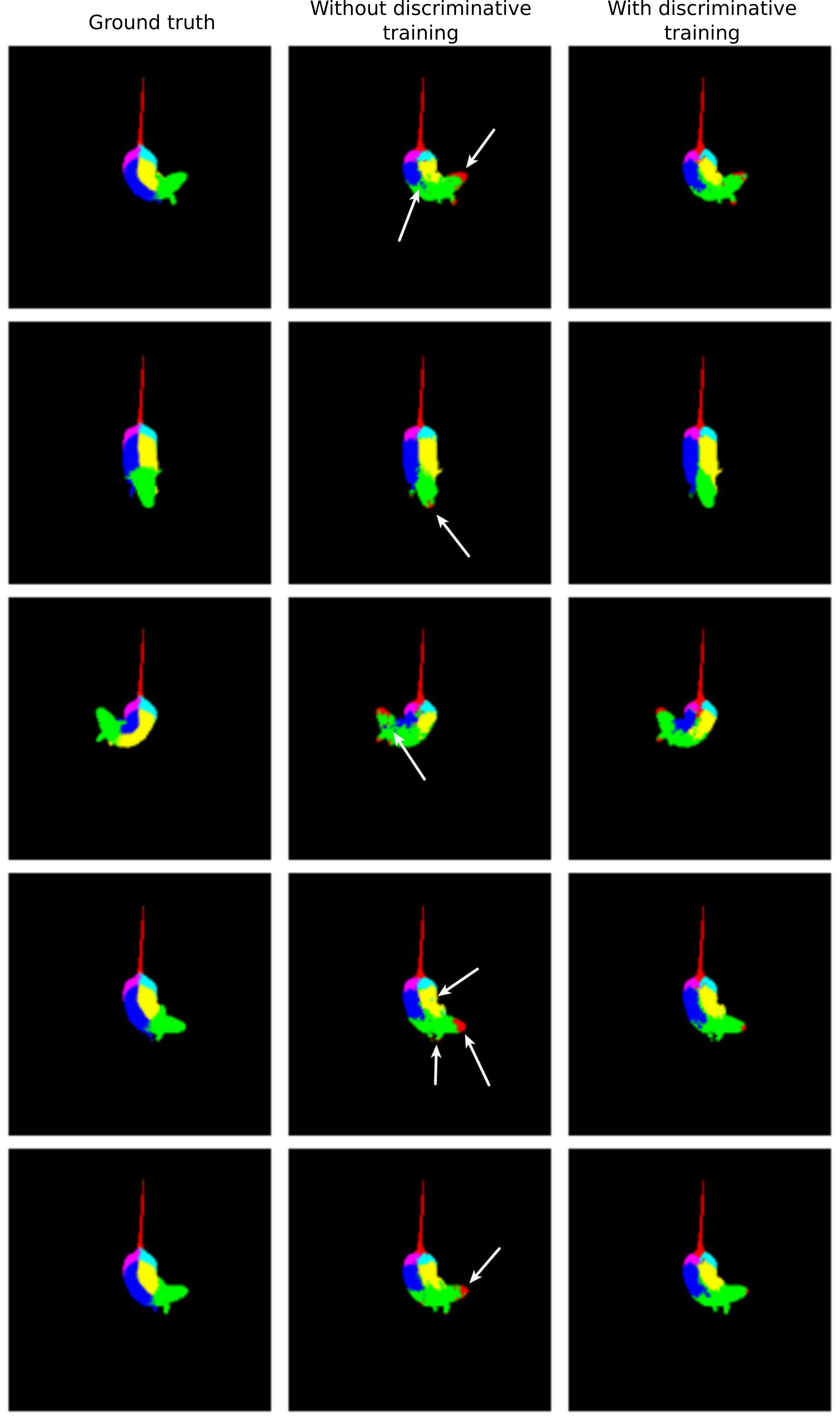}
\caption
{
    \footnotesize
    {
        Comparison of part-based labeling results using forests without and with discriminative training.
        Results demonstrate that with discriminative training, fewer pixels in head region of mouse are mislabled as tail (red color).
        It can also be observed that pixels in the middle of mouse body have fewer disconnected components.
    }
}
\label{fig:dis_train_pbl}
\end{figure*}

Fig.~\ref{fig:dis_train_pbl} shows several exemplar results comparing the part-based labeling results on synthetic data using forests without discriminative training ($m = 30$) and with discriminative training ($m = 100$).
These results can be compared to the ground-truth labeled image shown in leftmost column.
It can be observed that discriminately trained forest improves the results in a few problematic areas in the head region and middle region of mouse.

\begin{figure}[!tp]
\centering
\includegraphics[width=0.9\linewidth]{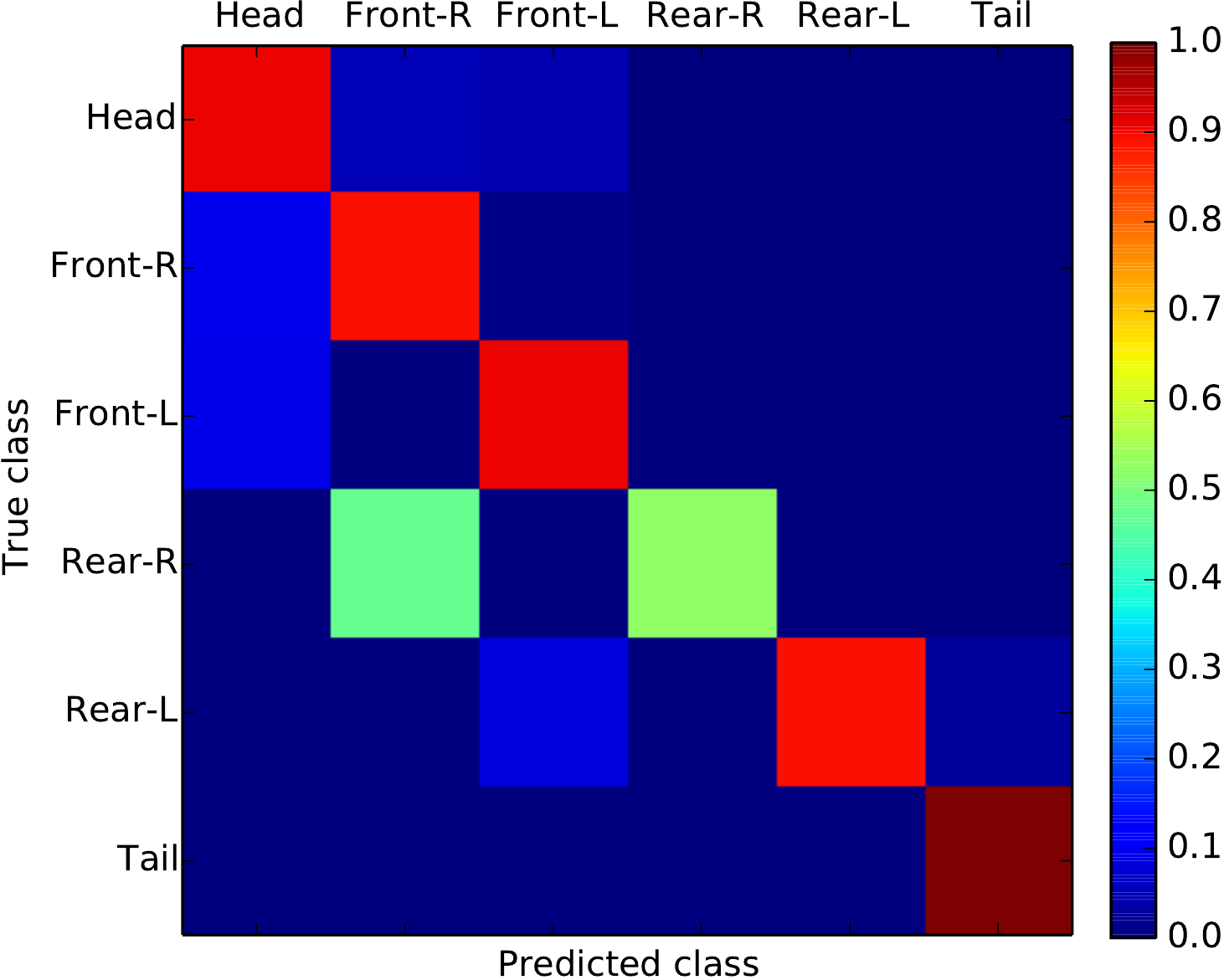}
\caption
{
    \footnotesize
    {
        Confusion matrix of our part-based labeling results on a mouse body partitioned into six classes.
    }
}
\label{fig:bpc-conf-matrix}
\end{figure}

Fig.\ref{fig:bpc-conf-matrix} presents the confusion matrix results for part-based labeling of single depth images with noise $\sigma = 16$ with the final forest.
All classes, except for the rear-right class, have an accuracy of $89\%$ or higher.
The rear-right class has an accuracy of $52.5\%$, as its pixels are sometimes misclassified as that from front-right.

\subsubsection{Real datasets}

\begin{figure}[!tp]
\centering
\subfigure[]
{
    \includegraphics[width=.45\textwidth]{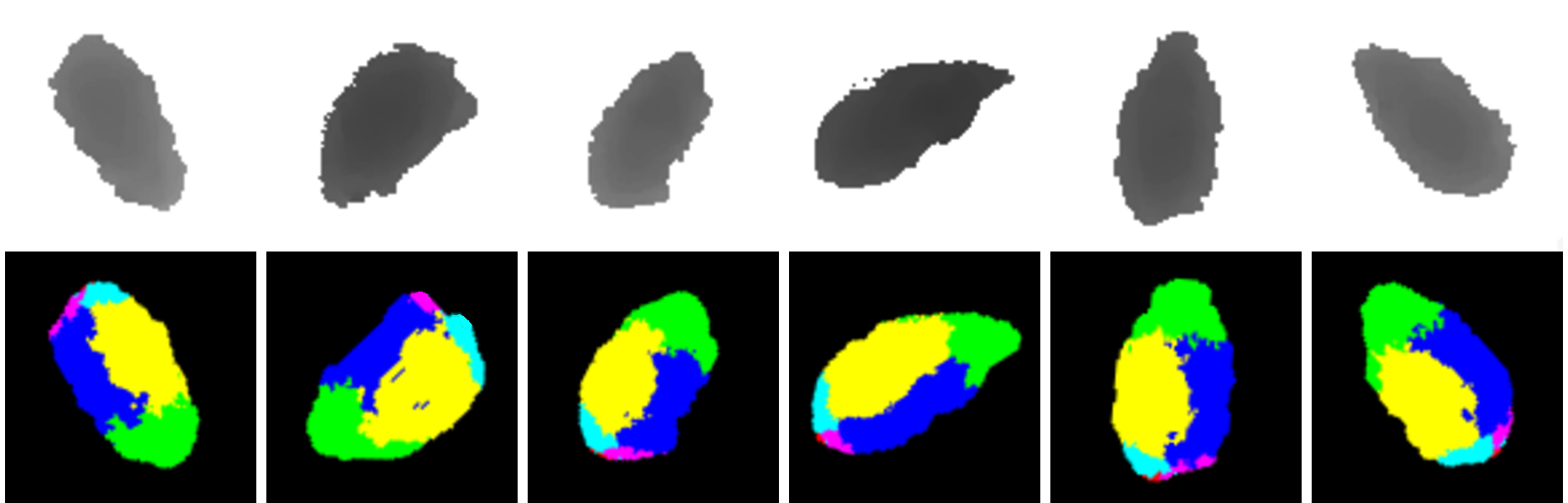}
    \label{fig:pbl-primesense}
}
\subfigure[]
{
    \includegraphics[width=.45\textwidth]{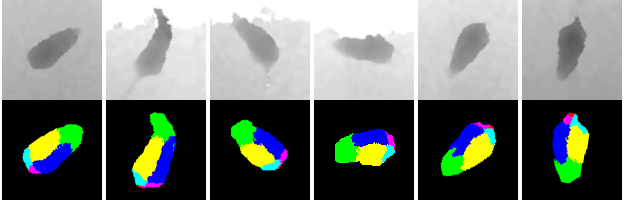}
    \label{fig:pbl-softkinetic}
}
\caption
{
    \footnotesize
    {
        Part-based labeling results on single depth images from (a) Primesense camera (b) SoftKinetic camera.
    }
}
\label{fig:pbl-real}
\end{figure}

To demonstrate the applicability of our method on real data, we perform part-based labeling on single depth images of a mouse captured from a top depth camera.
The real data results show a bit more noise, but is within expected bounds of part-based labeling for pose estimation as explained in in~\cite{ShoEtAl:cvpr11}.
To further demonstrate the versatility of the proposed approach, two types of depth cameras are considered.
For experiments, the depth images are preprocessed to remove shadows and noise and the patch containing mouse is recognized using contour detection.
Fig.~\ref{fig:pbl-primesense} and Fig.~\ref{fig:pbl-softkinetic} shows the results of our part-based labeling method on test data captured from a SoftKinetic and Primesense camera respectively.

Since the tail is hardly visible in depth images, our labeling usually misses the tail class.
Similar to the synthetic results, the rear right label region is often quite small comparing to the rest part labels.
Overall our approach is shown to be capable of providing satisfactory results robustly in this context.

\begin{figure}[!tp]
\centering
\subfigure[]
{
    \includegraphics[width=.23\textwidth]{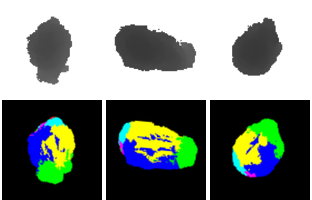}
    \label{fig:pbl-primesense-bad}
}
\subfigure[]
{
    \includegraphics[width=.23\textwidth]{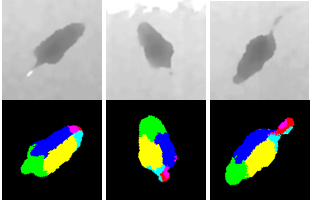}
    \label{fig:pbl-softkinetic-bad}
}
\caption
{
    \footnotesize
    {
    (a) Failed part based labeling results for depth images of Primesense camera.
    (b) Failed part-based labeling results for depth images of SoftKinetic camera.
    Some of the errors are mouse labels being reversed from head to tail.
    }
}
\end{figure}

Fig.~\ref{fig:pbl-primesense-bad} shows examples of failures in part-based labeling on images from Primesense camera.
Fig.~\ref{fig:pbl-softkinetic-bad} shows examples of failures in part-based labeling on images from SoftKinetic camera.
Other than tiny regions being mislabeled, another cause of failures is that sometimes the head is mistaken for the tail region.

\section{Conclusion and future work}

In this paper we presented a discriminative training approach that is demonstrated to reduce errors in both regression and classification random forests.
We also introduced an approach for mouse 3D pose estimation based mainly on single depth images.
The proposed approach is capable of producing reliable 3D full-body pose predictions when incorporating additional color image information from a different view. It also works well with various types of depth cameras.
We have also demonstrated satisfactory performance when addressing the related problem of part-based labeling.
Future work includes extensions to deal with mice tracking and group behavior analysis using depth cameras.

{\small
\bibliographystyle{spmpsci}      
\bibliography{main}
}

\end{document}